\documentclass[11pt, a4paper, logo, copyright, nonumbering]{xiaomi}
\usepackage[authoryear, compress, round]{natbib}
\usepackage{dblfloatfix}
\usepackage{ulem}
\usepackage{caption}
\usepackage{dramatist}
\usepackage{xspace}
\usepackage{pifont} %
\usepackage{multirow}
\usepackage{tcolorbox}
\usepackage{xltabular}
\usepackage{longtable}
\usepackage{hyperref}
\usepackage{wrapfig}
\usepackage{graphicx}

\usepackage{amsfonts}
\usepackage{amsmath}
\usepackage{amssymb}
\usepackage{lineno}
\usepackage{multirow}
\usepackage{adjustbox}

\usepackage[bottom]{footmisc}

\usepackage{CJKutf8}
\usepackage{subfigure}
\usepackage{setspace}
\usepackage{makecell}
\usepackage{graphicx}
\usepackage{subcaption}
\usepackage{multicol} %
\usepackage{siunitx}  %

\newcommand{\mimo}{MiMo-7B}
\newcommand{\mimovl}{MiMo-VL-7B}
\newcommand{\mimovlsft}{MiMo-VL-7B-SFT}
\newcommand{\mimovlrl}{MiMo-VL-7B-RL}

\newcommand{\mimobase}{MiMo-7B-Base}

\definecolor{xiaomiorange}{HTML}{FF6901}

\title{\centering MiMo-VL Technical Report}
\author{
 LLM-Core Xiaomi
}

\begin{abstract}
We open-source \mimovlsft{} and \mimovlrl{}, two powerful vision-language models delivering state-of-the-art performance in both general visual understanding and multimodal reasoning. 
\mimovlrl{} outperforms Qwen2.5-VL-7B on 35 out of 40 evaluated tasks, and scores 59.4 on OlympiadBench, surpassing models with up to 78B parameters.
For GUI grounding applications, it sets a new standard with 56.1 on OSWorld-G, even outperforming specialized models such as UI-TARS.
Our training combines four-stage pre-training (2.4 trillion tokens) with Mixed On-policy Reinforcement Learning (MORL) integrating diverse reward signals. We identify the importance of incorporating high-quality reasoning data with long Chain-of-Thought into pre-training stages, and the benefits of mixed RL despite challenges in simultaneous multi-domain optimization.
We also contribute a comprehensive evaluation suite covering 50+ tasks to promote reproducibility and advance the field.
The model checkpoints and full evaluation suite are available at \textcolor{xiaomiorange} {\url{https://github.com/XiaomiMiMo/MiMo-VL}}.
\end{abstract}

\begin{document}
\maketitle

\vspace{1.25cm}

\begin{figure}[htbp]
    \centering
    \includegraphics[width=0.54\textwidth]{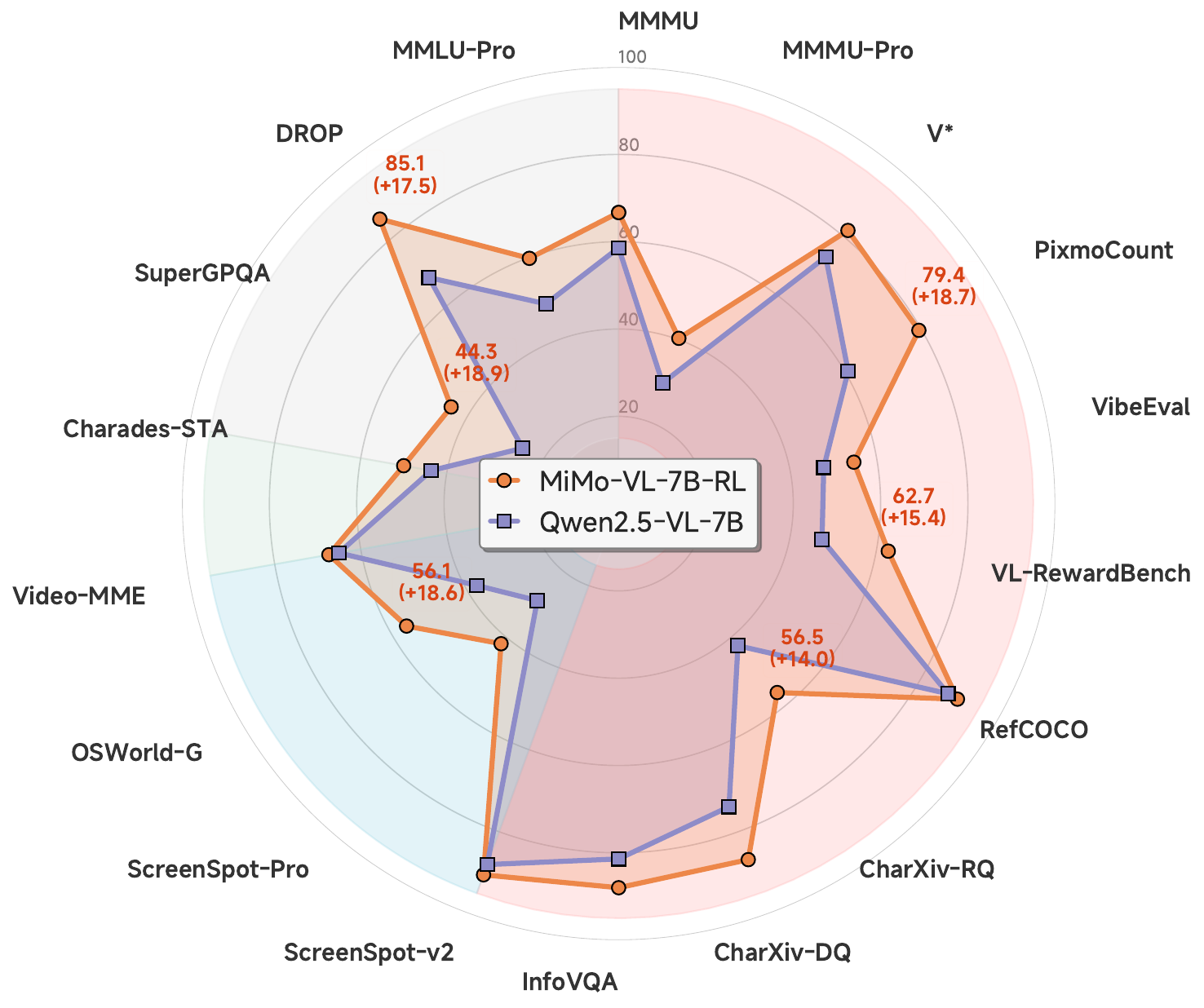}
    \hfill
    \includegraphics[width=0.45\textwidth]{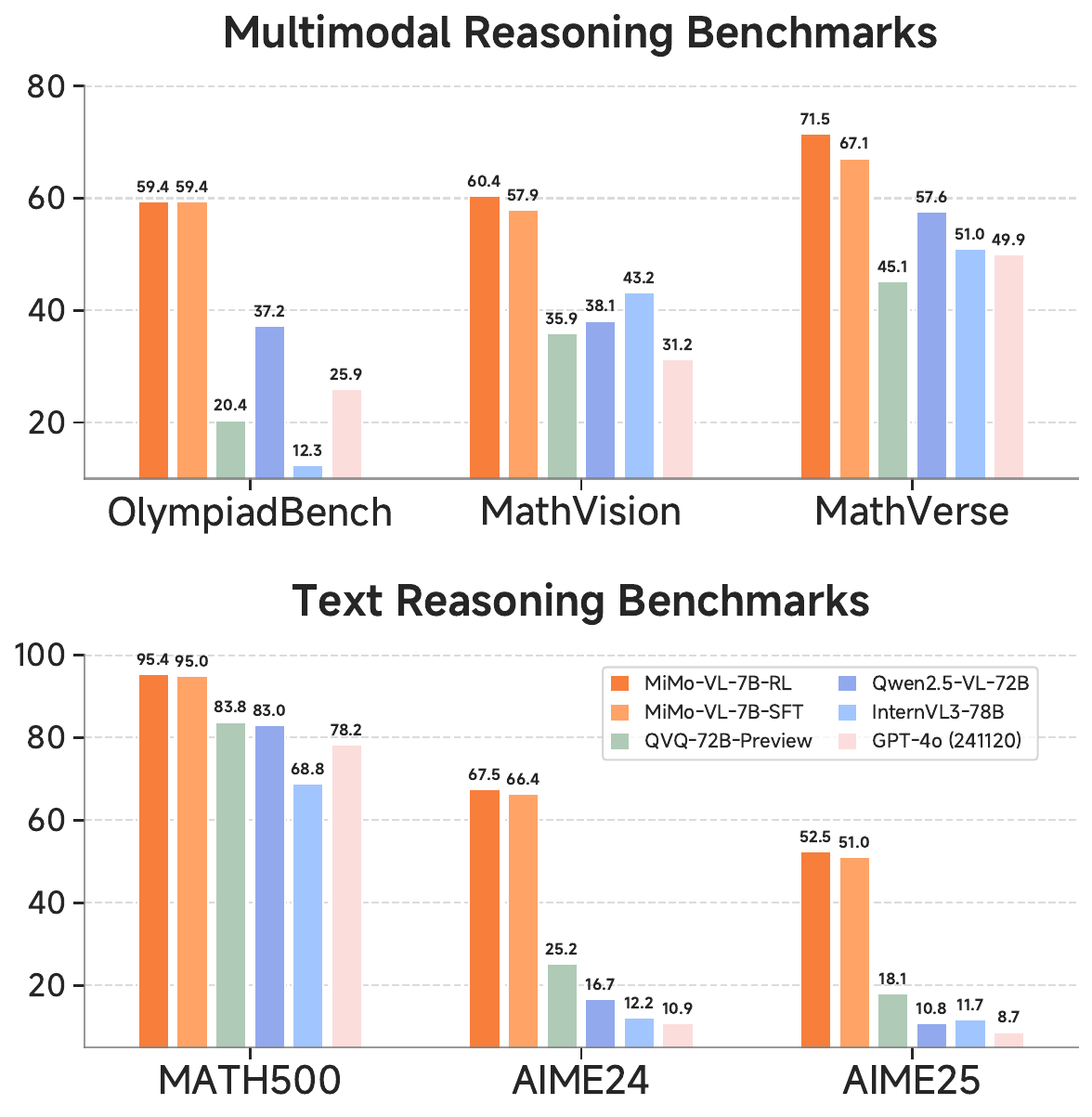}
    \caption{
    Benchmark performance of \mimovl{}.
    }
    \label{fig:mimo_comparison}
\end{figure}

\newpage

\begin{spacing}{0.9}
\tableofcontents
\end{spacing}

\newpage

\section{Introduction}
Vision-language models (VLMs) have emerged as the foundational backbone for multimodal AI systems, enabling autonomous agents to perceive the visual world, reason over multimodal content~\citep{yue2023mmmu}, and interact with both digital~\citep{xie2024osworld,cua2025} and physical environments~\citep{rt2,pi0}. 
The significance of these capabilities has motivated extensive exploration across multiple dimensions, including novel architectural designs~\citep{Alayrac2022FlamingoAV,team2024chameleon,dream2025} and innovative training methodologies with optimized data recipes~\citep{prismaticVLM,nvlm2024}, leading to rapid progress in the field~\citep{liu2023llava,tong2024cambrian,Qwen25VL}.

In this report, we share our efforts to build a compact yet powerful VLM, \mimovl{}. \mimovl{} comprises three components: (1) a native-resolution Vision Transformer (ViT) encoder that preserves fine-grained visual details, (2) a Multi-Layer Perceptron (MLP) projector for efficient cross-modal alignment, and (3) our \mimo{}~\citep{xia2025mimo} language model, specifically optimized for complex reasoning tasks. 

The development of \mimovl{} involves two sequential training processes: 
(1) A \emph{four-stage pre-training phase}, which includes projector warmup, vision-language alignment, general multimodal pre-training, and long-context Supervised Fine-Tuning (SFT). 
Throughout these stages, we curate high-quality datasets by strategically combining open-source collections with synthetic data generation techniques, consuming 2.4 trillion tokens, adjusting the dataset distribution in different stages to facilitate training. This phase yields the \textbf{\mimovlsft{}} model. 
(2) A subsequent \emph{post-training phase}, where we introduce Mixed On-policy Reinforcement Learning (MORL), a novel framework that seamlessly integrates diverse reward signals spanning perception accuracy, visual grounding precision, logical reasoning capabilities, and human preferences. 
We adopt the idea of GRPO~\citep{shao2024deepseekmath} and enhance training stability by exclusively performing on-policy gradient updates during this stage. This phase yields the \textbf{\mimovlrl{}} model.

During this journey, we find:

\textbf{(1) Incorporating high-quality, broad-coverage reasoning data from the pre-training stage is crucial for enhancing model performance. } 
In the current era of ``thinking'' models, vast quantities of multimodal pre-training data are undergoing a significant re-evaluation. Traditional question-answering (QA) data, with its direct, short answers, often restricts models to superficial pattern matching and leads to overfitting. 
In contrast, synthesized reasoning data with long Chain-of-Thought~(CoT) enables learning of complex logical relationships and generalizable reasoning patterns, providing richer supervision signals that markedly improve both performance and training efficiency.
To leverage this advantage, we curate high-quality reasoning data by identifying diverse queries, employing large reasoning models to regenerate responses with long CoT, and applying rejection sampling to ensure quality.
Furthermore, rather than treating this as supplementary fine-tuning data, we incorporate substantial volumes of this synthetic reasoning data directly into the later pre-training stages, where extended training yields continued performance improvements without saturation.

\textbf{(2) Mixed On-policy Reinforcement Learning further enhances model performance, while achieving stable simultaneous improvements remains challenging.}
We apply RL across diverse capabilities, including reasoning, perception, grounding, and human preference alignment, spanning modalities including text, images, and videos.
While this hybrid training approach further unlocks the model's potential, interference across data domains remains challenging.
Disparities in the growth trends of response length and task difficulty levels hinder stable, simultaneous improvements across all capabilities.

\mimovlrl{} delivers exceptional results across the full spectrum of multimodal capabilities.
\textbf{(1) On fundamental visual perception tasks,} it achieves state-of-the-art performance among open-source VLMs of comparable scale, scoring 66.7 on MMMU~\citep{yue2023mmmu} and outperforming Qwen2.5-VL-7B~\citep{Qwen25VL} on 35 out of 40 evaluated tasks. 
\textbf{(2) For complex multimodal reasoning,} \mimovlrl{} exhibits outstanding performance, scoring 59.4 on OlympiadBench~\citep{he2024olympiadbench} and surpassing models up to 72B parameters.
\textbf{(3) In GUI grounding for agentic applications,} our model sets a new standard by achieving a score of 54.7 on OSWorld-G~\citep{osworldg}, even exceeding specialized models like UI-TARS~\citep{qin2025uitars1}.
\textbf{(4) In terms of user experience and preference,} \mimovlrl{} achieves the highest Elo rating among all open-source VLMs on our in-house user preference evaluation, performing competitively with proprietary models such as Claude 3.7 Sonnet.

These results validate our approach: by combining strong perception, sophisticated reasoning, and precise grounding capabilities through our MORL framework, \mimovlsft{} and \mimovlrl{} establish new standards for open-source vision-language models.
To promote transparency and reproducibility, we also contribute a comprehensive evaluation suite covering 50+ tasks with complete prompts and protocols, enabling the community to build upon our work.

\section{Pre-Training}
In this section, we introduce the architecture design of our \mimovl, followed by the data curation process and training recipes of pre-training stages.

\subsection{Architecture}

\mimovl{} consists of three components: (1) a Vision Transformer (ViT) for encoding visual inputs such as images and videos; (2) a projector that maps the visual encodings into a latent space aligned with the Large Language Model (LLM); and (3) the LLM itself, which performs textual understanding and reasoning. 
To support native resolution inputs, we adopt Qwen2.5-ViT~\citep{Qwen25VL} as our visual encoder. 
We initialize the LLM backbone with \mimobase{}~\citep{xia2025mimo} to leverage its strong reasoning capability, and utilize a randomly initialized Multi-Layer Perceptron (MLP) as the projector. 
The overall architecture is illustrated in Figure~\ref{fig:model_arch}, and the model configuration can be found in Appendix~\ref{sec:model_config}.

\begin{figure}[t!]
    \centering
    \includegraphics[width=0.95\linewidth]{figures/mimo_vl_arch.pdf}
    \caption{Model architecture of \mimovl{}.}
    \label{fig:model_arch}
\end{figure}

\begin{table}[hbt]
\centering
\small
\resizebox{0.98\textwidth}{!}{
\begin{tabular}{@{}l|cccc@{}}
\toprule
\textbf{Stages} & \textbf{Stage 1} & \textbf{Stage 2} & \textbf{Stage 3} & \textbf{Stage 4} \\
\midrule
\multirow{2}{*}{\textbf{Purpose}} & 
Projector & 
Vision-Language & 
Multimodal & 
Long-context \\
& Warmup & Alignment & Pre-training & SFT \\ \midrule

\multirow{6}{*}{\textbf{Dataset}} 
&                      &                  &        +         & + \\
&                      &                  & Pure Text, & Long Pure Text, \\
&                      &        +         & OCR, Grounding, & Long Documents, \\
& Image-Caption Pairs  & Interleaved Data & QA, Video, GUI, & High-resolution Images, \\
&                      &                  & Instruction Data, & Extended Videos, \\
&                      &                  & Reasoning Data & Long Reasoning Data \\ \midrule

\textbf{Learning Rate} & 
1e-3 & 
1e-4, 1e-5 & %
1e-5 & 
2.5e-5 \\

\textbf{Training Tokens} & 
300B & 
167B & 
1.4T & 
550B \\

\textbf{Sequence Length} & 
8K & 
8K & 
8K & 
32K \\

\textbf{Trainable Components} &  Projector & ViT, Projector & All & All \\

\bottomrule
\end{tabular}
}
\caption{Overview of \mimovl{} training stages.}
\label{tab:training_stages}
\end{table}

\subsection{Pre-training Data}
The \mimovl{} pre-training dataset comprises 2.4 trillion tokens of high-quality, diverse multimodal data spanning images, videos, and text. This comprehensive dataset includes general image captions, interleaved data, Optical Character Recognition~(OCR) data, grounding data, video content, GUI interactions, reasoning examples, and text-only sequences.

To ensure the quality of each data modality, we implement dedicated data curation pipelines tailored to the characteristics of each data type. Throughout the training process, we systematically adjust the proportions of different data modalities across various training stages to optimize both training efficiency and model stability. Additionally, we employ phash-based image deduplication to eliminate potential overlaps between our training datasets and evaluation benchmarks, thereby minimizing data contamination.

We detail the specific processing procedures for each data type in the following sections.

\subsubsection{Image Caption Data} 
The construction of our image caption dataset involves a multi-stage process designed to ensure high quality and distributional balance. We begin by aggregating a substantial volume of publicly available caption data from web sources. This initial corpus then undergoes a rigorous deduplication phase, employing image perceptual hashing (phash) in conjunction with text filtering, to yield a refined set of unique raw captions.

Subsequently, leveraging both the images and their original textual descriptions as priors, we utilize a specialized captioning model to re-caption the entire raw caption dataset. Following re-captioning, we apply filtering mechanisms to the generated captions based on linguistic consistency and repetition patterns to ensure the quality of the re-captioned text. To address inherent data imbalances, we adopt the methodology of MetaCLIP~\citep{MetaCLIP}, which involves constructing novel bilingual (Chinese and English) metadata. This critical step serves to refine the caption distribution, thereby mitigating the over-representation of high-frequency entries and reducing dataset noise.

The culmination of this meticulous process is a balanced, high-quality, and diverse caption dataset. Crucially, we observe that this rich dataset significantly enhances model generalization and qualitative performance, offering advantages not always fully reflected in evaluations on existing, specialized benchmarks.

\subsubsection{Interleaved Data}
We compile an extensive corpus of interleaved image-text data from diverse sources, including webpages, books, and academic papers. For content originating from books and papers, we employ advanced PDF parsing toolkits for content extraction and cleansing. The filtering process prioritizes and retains data types rich in world knowledge, such as textbooks, encyclopedias, manuals, guides, patents, and biographies. Within this interleaved dataset, textual segments are evaluated based on metrics including knowledge density and readability. For the visual components, we implement filters to exclude images of diminutive size, anomalous aspect ratios, unsafe content, and those with minimal visual information (e.g., decorative chapter headings and illustrations). Finally, image-text pairs are scored on relevance, complementarity, and the balance of information density, ensuring the retention of high-quality data. This curated dataset significantly augments the model's knowledge repository, thereby establishing a robust foundation for its subsequent reasoning capabilities.

\subsubsection{OCR and Grounding Data}
To enhance the model's capabilities in OCR and object grounding, we compile an extensive corpus of OCR and grounding data from open-source datasets for pre-training. 

For the OCR data, the images encompass a diverse textual content from documents, tables, general scenes, product packaging, and mathematical formulas. To increase learning difficulty, in addition to standard printed text, we specifically incorporate images containing handwritten script, typographically deformed text, and blurry/occluded text, thereby improving the model's recognition performance and robustness. For a portion of this data, textual regions are annotated with bounding boxes, enabling the model to simultaneously predict these locations. 

For the grounding data, our images feature scenarios with both single and multiple objects. We further improve the model's capacity to comprehend complex referential intentions by employing intricate object expressions within localization prompts. In all scenarios involving localization, we use absolute coordinates for representation.

\subsubsection{Video Data}

Our video dataset primarily draws from publicly available online videos, spanning a wide array of domains, genres, and durations. Based on the videos, we design a video re-captioning pipeline that produces dense, fine-grained event-level descriptions. Each caption is temporally grounded with precise start and end timestamps, enabling the model to develop general video perception with time-awareness. From the caption dataset, we further collect a subset with a balanced distribution of event durations for temporal grounding pretraining. We also curate video analysis data that summarize the global semantics of videos, such as narrative structures, stylistic elements, and implicit intentions. These analytical paragraphs enable the model to grasp in-depth comprehension of videos and extended world knowledge. To enhance the model’s conversational coherence, we collect diverse and challenging questions about videos and synthesize corresponding responses. We also incorporate open-source video captioning and conversation datasets to further enrich our video pretraining data.

\subsubsection{Graphical User Interface Data}

To enhance the model's capabilities in navigating Graphical User Interfaces (GUIs), we collect open-source pre-training data covering all sorts of platforms such as mobile, web, and desktop. A synthetic data engine is also devised to compensate for the limitations of open-source data and to strengthen specific aspects of the model's capabilities. For example, we have constructed a vast amount of Chinese GUI data to enable the model to better handle Chinese GUI scenarios.

For GUI Grounding, we gather data for both element grounding and instruction grounding. Element grounding trains the model to precisely locate interface elements based on textual descriptions, establishing a robust perception of static user interfaces. Instruction grounding requires the model to identify target objects on screenshots according to user instructions, enhancing comprehension of GUI interaction logic. For this part, we additionally introduce a pre-training task that involves predicting intermediate actions based on before-and-after screenshots. Empirical evidence demonstrates that this approach significantly improves the model's dynamic perception of GUI interfaces.

For GUI Action, we collect a large scale of long GUI action trajectories. To ensure consistency across different platforms, we unified actions from mobile, web, and desktop environments into a standardized action space. A detailed specification of this action space is provided in Appendix~\ref{sec:gui_action_space}. This harmonization prevents action conflicts while maintaining platform-specific functionality.

\subsubsection{Synthetic Reasoning Data}
Our approach to generating synthetic reasoning data begins with the comprehensive curation of open-source questions. This diverse collection spans perceptual question answering, document question answering, video question answering, and visual reasoning tasks, supplemented by question-answer pairs derived from web content and literary works.

Following initial filtering of these source questions, we leverage a large reasoning model to produce answers that integrate explicit reasoning. A cornerstone of our methodology is rigorous, multi-stage quality control. Beyond verifying the factual correctness of answers, we apply strict filtering criteria to the reasoning processes themselves, evaluating clarity of thought, eliminating redundancy, and ensuring consistent formatting.

The resulting high-fidelity dataset plays a critical role in empowering our model. It facilitates the effective inheritance of strong reasoning abilities inherent in MiMo-7B-Base~\citep{xia2025mimo}, enabling their seamless transfer and adaptation to multimodal contexts. Consequently, this allows our model to exhibit powerful and versatile multimodal reasoning capabilities across a broad array of domains.

\subsection{Pre-training Stages}

Our model undergoes a four pre-training stages as illustrated in Table~\ref{tab:training_stages}:

\textbf{Stage 1}: We freeze the ViT and LLM components, and warm up the randomly initialized projector using image-caption pairs.
This ensures the projector learns to map visual concepts to the language model's representation space effectively, providing informative gradient signals for subsequent training stages rather than noisy updates from a poorly aligned projector.

\textbf{Stage 2}: The ViT is then unfrozen, and interleaved data is introduced to further strengthen vision-language alignment. The inclusion of complex and diverse images within the interleaved data enhances the ViT's performance and robustness.

\textbf{Stage 3}: In this stage, all parameters are trainable. We introduce a more diverse array of data and tasks, including OCR, grounding, video, and GUI data—accumulating to 1.4 trillion tokens—to bolster the model's general multimodal capabilities. To ensure stable mid-stage evaluation monitoring, small quantities of QA, instruction-following, and reasoning data are incorporated. Furthermore, a limited amount of text-only data is utilized to preserve \mimobase{}'s textual capability.

\textbf{Stage 4}: This stage is dedicated to enhancing the model's adaptability to long-context inputs. 
The training sequence length is extended from 8K to 32K tokens. 
We introduce additional data types such as long pure text, high-resolution images, long documents, extended videos, and long reasoning data to augment its long-context processing capabilities. 
As long-context packing significantly increases the effective batch size, the learning rate is adjusted from 1e-5 to 2.5e-5. 
Relative to Stage 3, this stage features a markedly increased proportion of reasoning data, alongside the introduction of long-form reasoning patterns.

These four stages create a powerful model, \mimovlsft{}. With particular emphasis on Stage 4, the model's reasoning capabilities are fully realized, enabling it to address highly intricate STEM problems. This advanced reasoning aptitude also generalizes effectively to common perception tasks. Consequently, our model demonstrates exceptionally high performance across various downstream benchmarks.

\section{Post-Training}
Building upon the visual perception capabilities and multimodal reasoning established during pre-training, we conduct post-training to further enhance \mimovl{}. 
Our approach employs a novel Mixed On-policy Reinforcement Learning~(MORL) framework that seamlessly integrates Reinforcement Learning with Verifiable Rewards (RLVR)~\citep{shao2024deepseekmath,lambert2025tulu3pushingfrontiers} with Reinforcement Learning from Human Feedback (RLHF)~\citep{ouyang2022instructgpt} to improve \mimovl{} on challenging reasoning tasks and alignment with human preferences.

\subsection{Reinforcement Learning with Verifiable Rewards}

RLVR relies exclusively on rule-based reward functions, enabling continuous model self-improvement. In the post-training of \mimovl{}, we design multiple verifiable reasoning and perception tasks where final solutions can be precisely validated using predefined rules.

\paragraph{Visual Reasoning} 
Visual reasoning capabilities are fundamental for multimodal models to understand and solve complex problems that require both visual perception and logical thinking. To facilitate this capability, we compile diverse verifiable STEM questions from open-source communities and proprietary K-12 collections. 
An LLM is prompted to filter proof-based problems and rewrite multiple-choice questions with numerical or symbolic answers into free-answer formats, alleviating potential reward hacking.
We further refine question quality through comprehensive model-based difficulty assessment, excluding questions that either cannot be solved by advanced VLMs or are too easy, with a \mimovl{} rollout pass rate exceeding 90\%.
Additionally, we remove questions solvable even without image inputs. After data cleaning and category balancing, we curate a visual reasoning dataset of 80K problems. For evaluation, we use the rule-based Math-Verify library to determine response correctness.\footnote{\url{https://github.com/huggingface/Math-Verify}}

\paragraph{Text Reasoning}
Since most visual reasoning data is limited to K-12 level questions, the reasoning performance of RL-trained models could be constrained.
In contrast, text-only reasoning datasets include more challenging queries requiring college or competition-level intelligence.
To unleash the full reasoning potential of our model, we incorporate mathematical reasoning data from \citet{xia2025mimo}. Rewards are computed using the same rule-based Math-Verify library to ensure consistent evaluation across both visual and textual reasoning tasks.

\paragraph{Image Grounding}
Accurate spatial localization is essential for models to understand object relationships and spatial reasoning in images. We include both general and GUI grounding tasks in our RLVR framework to enhance \mimovl{}'s grounding capability. For bounding box predictions, rewards are calculated using the Generalized Intersection over Union (GIoU)~\citep{rezatofighi2019generalized} between predicted and ground-truth boxes. For point-style outputs, rewards are determined by whether the predicted point falls within the ground-truth bounding box.

\paragraph{Visual Counting}
Precise counting abilities are essential for quantitative visual understanding and mathematical reasoning in visual contexts~\citep{chen2025r1v}. 
We enhance visual counting capabilities through RL training, where rewards are defined by the accuracy of the model's counting predictions compared to ground-truth counts.

\paragraph{Temporal Video Grounding}
Beyond static image understanding and reasoning, we extend our RLVR framework to dynamic video content to capture temporal dependencies. We incorporate temporal video grounding tasks that require the model to localize video segments corresponding to natural language queries~\citep{wang2025timezero}. The model outputs timestamps in \texttt{[mm:ss,mm:ss]} format to indicate the start and end times of the target video segments. Rewards are computed as the Intersection over Union (IoU) between predicted and ground-truth temporal segments.

\subsection{Reinforcement Learning from Human Feedback}
To align model outputs with human preferences and mitigate undesirable behaviors, we employ Reinforcement Learning from Human Feedback (RLHF) as a complementary approach to our verifiable reward framework.

\paragraph{Query Collection}

Query diversity is paramount to the success of RLHF. Our methodology commences with gathering multimodal and text-only queries from open-source instruction tuning datasets and in-house human-written sources. All collected queries, both text and multimodal, then undergo a dedicated screening process. To further enhance diversity, we employ techniques such as clustering queries based on their embeddings and analyzing the resultant patterns. Crucially, we balance the proportions of Chinese and English queries, as well as those targeting helpfulness and harmlessness, before curating the final query set. For each selected query, MiMo-VL-7B and multiple other top-performing VLMs are prompted to generate responses. These responses are subsequently pairwise ranked by an advanced VLM to construct the definitive dataset for reward model training. Notably, to mitigate potential reward hacking, this same query set is utilized for both reward model training and the RLHF process.

\paragraph{Reward Model}
We develop two specialized reward models tailored to different input modalities, training them using the Bradley-Terry reward modeling objective~\citep{ouyang2022instructgpt}. 
The text-only reward model is initialized from \mimo{}~\citep{xia2025mimo} to leverage its strong language understanding capabilities, while the multimodal reward model builds upon \mimovl{} to effectively process queries containing visual inputs. This dual-model approach ensures optimal performance across both textual and multimodal evaluation scenarios.

\subsection{Mixed On-Policy Reinforcement Learning}

\begin{figure}[t!]
    \centering
    \includegraphics[width=0.97\linewidth]{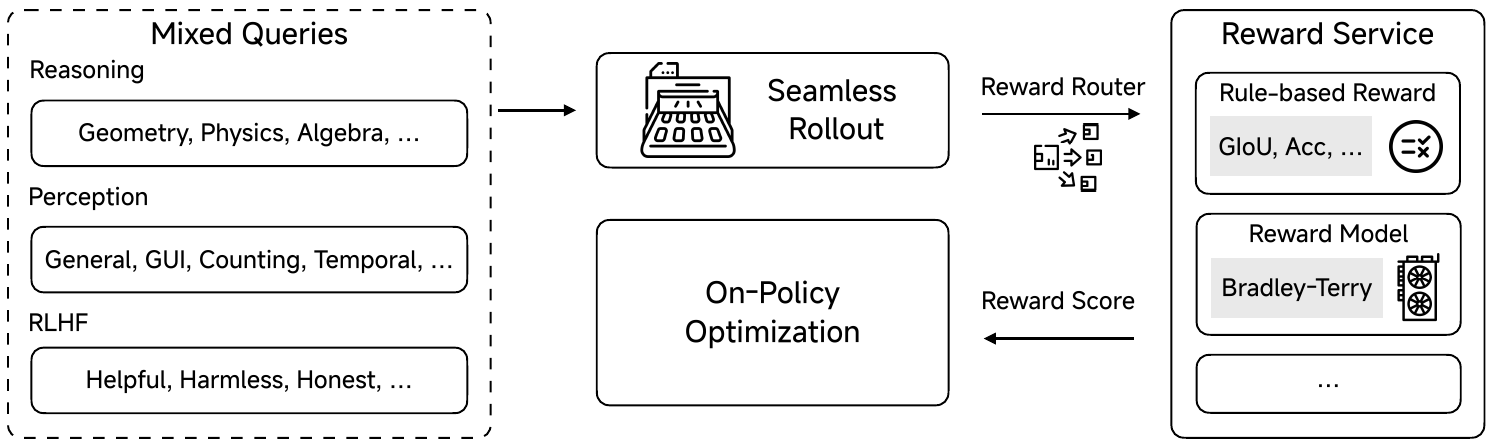}
    \caption{Mixed On-policy Reinforcement Learning in post-training phase.}
    \label{fig:morl_pipeline}
\end{figure}

In the post-training phase of \mimovl{}, we implement Mixed On-policy Reinforcement Learning (MORL) to simultaneously optimize RLVR and RLHF objectives.
As illustrated in Figure~\ref{fig:morl_pipeline}, we integrate rule-based and model-based rewards as unified services within the verl framework~\citep{sheng2024hybridflow}, enhanced by the Seamless Rollout Engine~\citep{xia2025mimo}.

\paragraph{On-Policy RL Recipe}
We adopt a fully on-policy variant of GRPO~\citep{shao2024deepseekmath} as the RL algorithm, which demonstrates robust training stability and effective exploration capabilities~\citep{chen2025acereason}.
For each problem $q$, the algorithm samples a group of responses $\left\{o_1,o_2,...,o_G\right\}$ from the policy $\pi_\theta$, and updates the policy by maximizing the following objective:
\begin{equation}
\label{eq:grpo}
    \mathcal{J}_{\mathrm{GRPO}}\left(\theta\right)= \mathbb{E}_{q\sim D,\{o_i\}_{i=1}^G\sim \pi_\theta(\cdot|q)} \left[\frac{1}{\sum_{i=1}^G\left|o_i\right|}\sum_{i=1}^G \sum_{j=1}^{\left|o_i\right|} A_{i,j}\right],
\end{equation}
where $A_{i,j}$ is the advantage, which is computed by the rewards $\left\{r_1,r_2,...,r_G\right\}$ of responses in the same group:
\begin{equation}
    A_{i,j} = \frac{r_i-\mathrm{mean}(\{r_i\}_{i=1}^G)}{\mathrm{std}(\{r_i\}_{i=1}^G)}.
\end{equation}
Compared to vanilla GRPO, this on-policy variant performs single-step policy updates following response rollout, eliminating the need for a clipped surrogate training objective.
Following \citet{xia2025mimo}, we integrate several advancements, including removal of the KL loss, dynamic sampling, easy data filter, and re-sampling strategies, into our RL training recipe.

\paragraph{Reward-as-a-Service}
The MORL process integrates tasks across reasoning, perception, grounding, multimodal RLHF, and text-only RLHF, each requiring distinct reward functions or dedicated reward models. 
To provide a unified interface and near-zero latency reward computation, we introduce Reward-as-a-Service (RaaS).
A reward router dynamically selects the appropriate reward function based on the query's task type.
To minimize latency, reward models are deployed as standalone services, ensuring scalable reward computation accessible via HTTP.
All rewards are normalized to the range $[0,1]$.
No additional reward, such as format rewards, is incorporated in our training process.

\section{Evaluation}
We evaluate \mimovl{} across 50 tasks to comprehensively assess its capabilities. Appendix~\ref{sec:eval_bench} lists all the benchmarks adopted in our evaluation. 
We also assess model performance with an in-house evaluation set.

\begin{table}[htbp]
    \centering
    \small
    \resizebox{0.99\textwidth}{!}{
    \begin{tabular}{l c | S[table-format=2.1] S[table-format=2.1] | S[table-format=2.1] S[table-format=2.1] S[table-format=2.1] | S[table-format=2.1] S[table-format=2.1]}
    \toprule
    \multirow{2}{*}{\textbf{Benchmark}}  &  \multirow{2}{*}{\textbf{Metrics}} & \textbf{MiMo-VL} & \textbf{MiMo-VL} & \textbf{Qwen2.5-VL} & \textbf{InternVL3} & \textbf{Gemma-3}  & \multicolumn{1}{c}{\multirow{2}{*}{\textbf{GPT-4o}}} & \multicolumn{1}{c}{{\textbf{Claude 3.7}}} \\
    & & \textbf{7B-SFT} & \textbf{7B-RL} & \textbf{7B} & \textbf{8B} & \textbf{27B-IT} &   & \textbf{Sonnet} \\
    \midrule
    \rowcolor{xiaomiorange!20}
    \multicolumn{9}{l}{\textbf{General}} \\ %
    MMMU$_{\mathrm{val}}^{\dag}$          & Acc.         & 64.6  & \textbf{66.7} & 58.6  & 62.7  & 64.9  &   70.7   & 69.8*                \\
    MMMU-Pro$_{\mathrm{standard}}$        & Acc.         & 45.2  & \textbf{46.2} & 34.7* & 45.6* & 37.8* &  42.5*   &  56.5*               \\
    MMMU-Pro$_{\mathrm{vision}}$          & Acc.         & 39.4  & \textbf{40.3} & 29.4* & 37.8* & 24.9* &    36.1*    &  45.8*               \\
    MMBench-en$_{\mathrm{test}}$          &  Acc.        &     \textbf{84.5}  & 84.4 & 83.5  & 83.4  &    81.6*   &    84.6*    &  84.8*               \\
    MMBench-cn$_{\mathrm{test}}$          &  Acc.        &      81.9 & 82.0  & \textbf{83.4} & 82.2  &    82.4*   &    84.5*    &  83.7*               \\
    Mantis                                &  Acc.        & \textbf{78.8} & 78.3  & 74.7* & 72.8* &    70.0*   &  75.6* &  75.1*               \\
    MME-RealWorld$_{\mathrm{en}}$         &  Acc.        & 57.4  & \textbf{59.1} & 57.4  & 56.1* &   51.9*    &    57.5*    &  50.8*               \\
    MME-RealWorld$_{\mathrm{cn}}$         &  Acc.        & 55.0  & 55.5  & 51.2* & \textbf{58.5}* &  47.9*     &    58.5*   &  40.6*               \\
    AI2D                                  &  Acc.        &  83.2 & 83.5 & 83.9  & \textbf{85.2}  & 84.5  &  82.6* &  81.4*               \\ 
    BLINK$_{\mathrm{val}}$                &  Acc.        &  \textbf{62.5} & 62.4 & 56.4  & 55.5  & 53.3* &  60.0  &  62.3*               \\ 
    CV-Bench                              & Acc.         &  81.8 & \textbf{82.3} & 75.4* & 81.0* & 70.4* &  76.0* & 75.4* \\ 
    VibeEval$^{\dag}$                      &  GPT-Score   & 47.2  & \textbf{54.7} & 47.7* & 43.6* & 44.0* & 64.7  &  39.0*               \\ 
    VL-RewardBench$^{\dag}$                &  Macro Acc.  &  61.9 & \textbf{62.7} & 47.3* & 49.7* & 51.9* &  62.4  & 67.4*                \\ 
    V* &  Acc.        &  80.6 & \textbf{81.7} & 73.8* & 72.8* & 50.8* &  73.9  &  {-}               \\  
    VLMs are Blind                        &  Acc.        & 78.0  & \textbf{79.4} & 37.4* & 36.8* &  18.6*     &   49.8*   &  72.1*               \\ 
    PixmoCount                            &  Acc.        &  \textbf{79.4} & \textbf{79.4} & 60.7* & 62.0  &   48.6*    &    54.4*  &  53.5*               \\ 
    CountBench                            &  Acc.        & 87.0  & \textbf{90.4} & 74.1* & 80.0* & 77.2*  &   85.7* &  90.2*               \\ 
    RefCOCO$_{\mathrm{val}}^\mathrm{avg}$              &  Acc.@0.5    & 85.7  & 89.6  & 87.1  & \textbf{90.1} &   {-}    &  {-}      &   {-}              \\  
    \midrule
    \rowcolor{xiaomiorange!20}
    \multicolumn{9}{l}{\textbf{Doc \& OCR}} \\ %
    ChartQA$^{\dag}$                      &  Acc.        & \textbf{92.9} & 91.7  & 90.2* & 89.6* & 78.0  &  86.7  & 92.2*                 \\
    CharXiv$_{\mathrm{RQ}}^{\dag}$        &  Acc.        & 54.4  & \textbf{56.5} & 42.5  & 37.6  &  29.2*     &  52.0 &   63.0*              \\
    CharXiv$_{\mathrm{DQ}}^{\dag}$        &  Acc.        &  \textbf{87.0} & 86.8  & 73.9  & 73.6  &   63.8*    &  86.5  &   89.5*              \\
    DocVQA$_{\mathrm{val}}^{\dag}$        &  Acc.        &  95.2 &   \textbf{95.7}    & 95.5* & 89.4* & 86.6  &    93.0*    &   94.1*              \\
    InfoVQA$_{\mathrm{val}}^{\dag}$       &  Acc.        & 87.2  & \textbf{88.0} & 81.4* & 70.7* & 70.6  &  82.1*  &    65.5*             \\
    SEED-Bench-2-Plus                     &  Acc.        & 71.9  & \textbf{72.4} & 70.7  & 69.7  &   66.3*    &   71.1*  & 72.9*                \\
    OCRBench$^{\dag}$                     &  Acc.        & 87.6  & 86.6  & \textbf{89.7}* & 88.0  &   77.6*    & 84.3*  &  80.6*               \\
    \midrule
    \rowcolor{xiaomiorange!20}
    \multicolumn{9}{l}{\textbf{GUI}} \\ %
    VisualWebBench$_{\mathrm{avg}}$       &  Acc.        &  \textbf{80.2} & \textbf{80.2}  & 72.9* & 62.7  &    49.7*   & 80.2  &  79.3*               \\
    WebSrc$_{\mathrm{val}}$               &  SQuAD F1    &  \textbf{96.5} & 95.3  & 94.6* & 91.1  &    89.0*   & 89.1*   & 91.1*                \\
    ScreenSpot                            & Center Acc.  &  \textbf{87.3} & 87.2  & 84.7  & 79.5  &   {-}    & 18.3   &  {-}               \\
    ScreenSpot-v2                         & Center Acc.  & 89.5  & \textbf{90.5} & 88.0  & 81.4  &   {-}    & 18.5   &  {-}               \\
    ScreenSpot-Pro$_{\mathrm{avg}}$                & Center Acc.  & 39.9  & \textbf{41.9} & 29.0  &  {-}  &   {-}    & {-}    &   {-}              \\
    OSWorld-G$_{\mathrm{no\_refusal}}$    & Center Acc.  &  54.7 & \textbf{56.1} & 37.5* &  {-}  &   {-}    & {-}    &   {-}              \\
    \midrule
    \rowcolor{xiaomiorange!20}
    \multicolumn{9}{l}{\textbf{Video}} \\ %
    Video-MME$_{\mathrm{w/o~sub.}}$                             & Acc.         & 66.9  & \textbf{67.4} & 65.1 & 66.3  &    {-}   & {-}    &     {-}            \\
    Video-MMMU                            & Acc.         &  \textbf{53.1} &   43.3   & 47.4  &   48.9*    &   {-}    & {-}    &    {-}             \\
    EgoSchema$_{\mathrm{val}}$                             & Acc.         & 60.4  & 59.6  & 62.4* &   \textbf{68.2}*   &   {-}   & {-}    &    {-}             \\
    Charades-STA                          & mIoU         &   38.5    & \textbf{50.0} & 43.6  &   25.4    &    {-}   & {-}    &  {-}               \\
    \midrule
    \rowcolor{xiaomiorange!20}
    \multicolumn{9}{l}{\textbf{Text}} \\ %
    GPQA Diamond                          & Pass@1       & 56.3 & \textbf{58.3} & 30.3* & 33.5* & 40.9  & 50.6* & 66.0* \\
    SuperGPQA                             & Pass@1       & 42.6 & \textbf{44.3}      & 25.4* & 29.4* & 29.3* & 31.6* & 54.4* \\
    DROP                                  & 3-shot F1    & 82.7 & \textbf{85.1} & 67.6* & 77.5* & 72.2  & 81.5  & 87.2* \\
    MMLU-Pro                              & EM           & 59.8 &  \textbf{64.8}     & 48.7* & 58.3* & 45.3  & 69.1  & 81.0  \\
    IF-Eval                               & Prompt Strict& 75.3 &   75.9    & 75.1* & 87.2* & \textbf{88.9} & 82.5  & 88.7* \\
    \bottomrule
    \end{tabular}
}
    \caption{
        Comparison of \mimovl{} models with other models on diverse visual-language and text benchmarks.
        Results marked with * are obtained using our evaluation framework. $^{\dag}$ indicates that evaluation is performed using GPT-4o. The best results among open-source models are \textbf{bolded}.
    }
    \label{tab:general_updated} %
\end{table}

\subsection{Evaluation Setting}

For image understanding benchmarks, we set the max pixels for the input image to 4096 × 28 × 28 and the maximum generation tokens to 32,768, employing greedy search for decoding. 
For video benchmarks, videos are sampled at 2 FPS, with a maximum of 256 frames and a total token limit of 16,384.
For text evaluation benchmarks, we set the max new tokens for generation as 32,768, temperature as 0.6 and top-p as 0.95.
We adapt the existing framework based on LMMs-Eval~\citep{lmmseval} to better accommodate long-CoT reasoning models. 
We further optimize the evaluation logic for specific tasks to ensure better evaluation consistency. 
To facilitate open research, we open-source our evaluation framework with all prompts used.\footnote{\url{https://github.com/XiaomiMiMo/lmms-eval}}

\subsection{General Capabilities}

Table~\ref{tab:general_updated} presents benchmark results assessing the general capabilities of VLMs. Our \mimovl{} models demonstrate consistently leading performance across a diverse range of vision-language and text benchmarks, establishing new state-of-the-art results among open-source models and even surpassing proprietary counterparts.

Specifically, our \mimovl{} models achieve state-of-the-art results among open-source models. \mimovlsft{} and
\textbf{(i)} On general vision-language tasks, our models achieve exceptional performance that leads the open-source field. \mimovlsft{} and \mimovlrl{} obtain 64.6\% and 66.7\% on MMMU$_{\mathrm{val}}$ respectively, outperforming much larger models such as Gemma 3 27B. For document and chart understanding, \mimovlrl{} excels with a top open-source score of 56.5\% on CharXiv$_{\mathrm{RQ}}$, significantly exceeding Qwen2.5-VL (42.5\%) by 14.0 points and InternVL3 (37.6\%) by 18.9 points.
\textbf{(ii)} Our models demonstrate superior video understanding capabilities while maintaining strong textual performance. 
MiMo-VL-SFT achieves a leading 53.1\% on Video-MMMU, and MiMo-VL-RL obtains an impressive 50.0\% mIoU on Charades-STA. \textbf{(iii)} Compared with \mimo{}, our models maintain decent performance on text-only benchmarks.
\textbf{(iv)} Remarkably, our MoRL yields comprehensive improvements, with the most impressive gains observed on challenging benchmarks such as VibeEval and CountBench.

\begin{table}[t]
    \centering
    \small
    \sisetup{table-align-text-post=false} %
    \resizebox{\textwidth}{!}{%
    \begin{tabular}{l c | S[table-format=2.1] S[table-format=2.1] | S[table-format=2.1] S[table-format=2.1] S[table-format=2.1] S[table-format=2.1] | S[table-format=2.1] S[table-format=2.1]}
    \toprule
    \textbf{Reasoning}  &  \multirow{2}{*}{\textbf{Metrics}} & {\textbf{MiMo-VL}} & {\textbf{MiMo-VL}} & {\textbf{QVQ-72B}} & {\textbf{Qwen2.5-VL}} & {\textbf{Intern-VL3}} & {\textbf{Qwen2.5}}  & \multicolumn{1}{c}{\multirow{2}{*}{\textbf{GPT-4o}}} & {\textbf{Gemini-2.5}} \\
    \textbf{Benchmark} & & {\textbf{7B-SFT}} & {\textbf{7B-RL}} & {\textbf{Preview}} & {\textbf{72B}} & {\textbf{78B}} & {\textbf{72B}} &  & {\textbf{Pro}} \\
    \midrule
    \rowcolor{xiaomiorange!20}
    \multicolumn{10}{l}{\textbf{Multi-modal}} \\ %
    OlympiadBench  & Acc.  & \textbf{59.4} & \textbf{59.4} & 20.4  & 37.2{*}  & 12.3 & {{-}} &  25.9 & 69.8   \\  
    MathVision     &  Acc. & 57.9 & \textbf{60.4} & 35.9 & 38.1    & 43.2 & {{-}} & 31.2  & 69.1    \\  
    MathVerse$_{\mathrm{vision\_only}}^{\dag}$  & Acc. & 67.1 & \textbf{71.5} & 45.1{*} & 57.6 & 51.0 & {{-}} & 49.9  &  76.7    \\  
    DynaMath         & Worst-case Acc. & \textbf{46.9}  & 45.9  & 30.7  & 38.1{*} & 35.1 & {{-}} & 48.5 & 56.3   \\
    WeMath & Strict Score & 65.1 & \textbf{66.3} & 37.7{*}  & 50.6{*} & 46.1& {{-}} & 50.6  & 78.0   \\
    LogicVista & Acc. & 61.2 & \textbf{61.4} & 53.8{*}  & 57.1{*}  & 55.9 & {{-}} & 64.4 & 73.8   \\
    MathVista$_{\mathrm{mini}}^{\dag}$ &  Acc. & \textbf{81.8} & 81.5 & 71.4  & 74.8 & 72.2 & {{-}} & 63.8  & 80.9    \\
    \midrule
    \rowcolor{xiaomiorange!20}
    \multicolumn{10}{l}{\textbf{Text}} \\ %
    MATH500 & Pass@1         & 95.0& \textbf{95.4} & 83.8{*} & 83.0 & 68.8{*} & 82.8{*} & 78.2{*} & 95.2 \\
    AIME24 & Pass@1   & 66.4 & \textbf{67.5} & 25.2{*} & 16.7{*} & 12.2{*} & 19.4{*} & 10.9{*} & 92.0  \\
    AIME25 & Pass@1   & 50.9 & \textbf{52.5} & 18.1{*} & 10.8{*} & 11.7{*} & 13.3{*} & 8.7{*}  & 86.7  \\
    \bottomrule
    \end{tabular}%
    }
    \caption{
        Comparison of \mimovl{} with other models on reasoning benchmarks.
        Results marked with * are obtained using our evaluation framework. $^{\dag}$ indicates that evaluation is performed using GPT-4o. The best results among open-source models are \textbf{bolded}.
    }
    \label{tab:reason_modified_final} %
\end{table}

\subsection{Reasoning Tasks}

Table~\ref{tab:reason_modified_final} presents evaluation results for multimodal and text reasoning benchmarks.
In multimodal reasoning, both the SFT and RL models significantly outperform all compared open-source baselines across these benchmarks.
Notably, \mimovlsft{} surpasses much larger models, including Qwen2.5-VL-72B and QVQ-72B-Preview.
The RL model further improves performance on most reasoning benchmarks.
For example, \mimovlrl{} boosts accuracy on MathVision from 57.9\% to 60.4\%.
\mimovl{} models also exhibit impressive reasoning capabilities on pure text benchmarks, even outperforming Qwen2.5-72B.
These results demonstrate that our multimodal pre-training and post-training recipes effectively endow the model with exceptional visual capabilities and strong text intelligence.

\subsection{GUI Tasks}

\begin{figure}[t]
    \centering
    \includegraphics[width=0.98\linewidth]{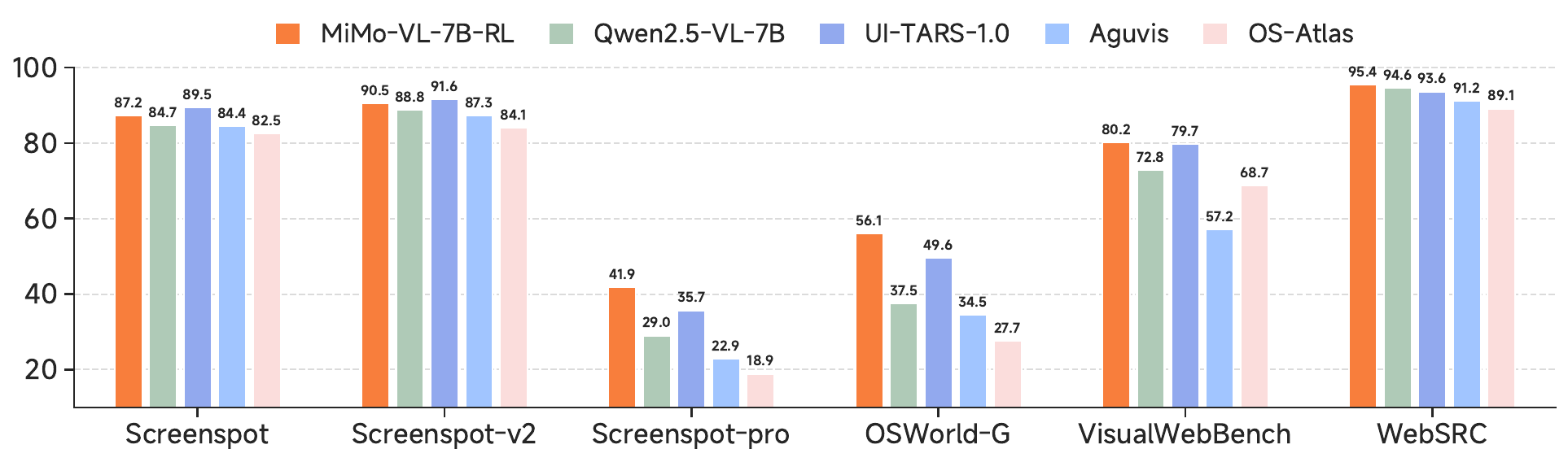}
    \caption{GUI understanding and grounding results. \mimovlrl{} achieves comparable results with GUI specialized models.}
    \label{fig:gui_specific}
\end{figure}

In addition, we demonstrate that \mimovl{} models possess exceptional GUI understanding and grounding capabilities. 
In Table~\ref{tab:general_updated}, \mimovlrl{} outperforms all other general VLMs compared. In Figure~\ref{fig:gui_specific}, we further compare \mimovlrl{} with GUI-specialized models (UI-TARS-1.0~\citep{qin2025ui}, Aguvis~\citep{xu2024aguvis}, OS-Atlas~\citep{wu2024atlas}) of similar size on GUI Understanding (WebSrc, VisualWebBench) and Grounding (Screenspot, Screenspot-v2, Screenspot-Pro, OSWorld-G) benchmarks. As a general-purpose VLM, MiMo-VL achieves comparable or even superior performance to GUI-specialized models, particularly on the more challenging Screenspot-Pro and OSWorld-G benchmarks.

\begin{figure}
    \centering
    \includegraphics[width=\linewidth]{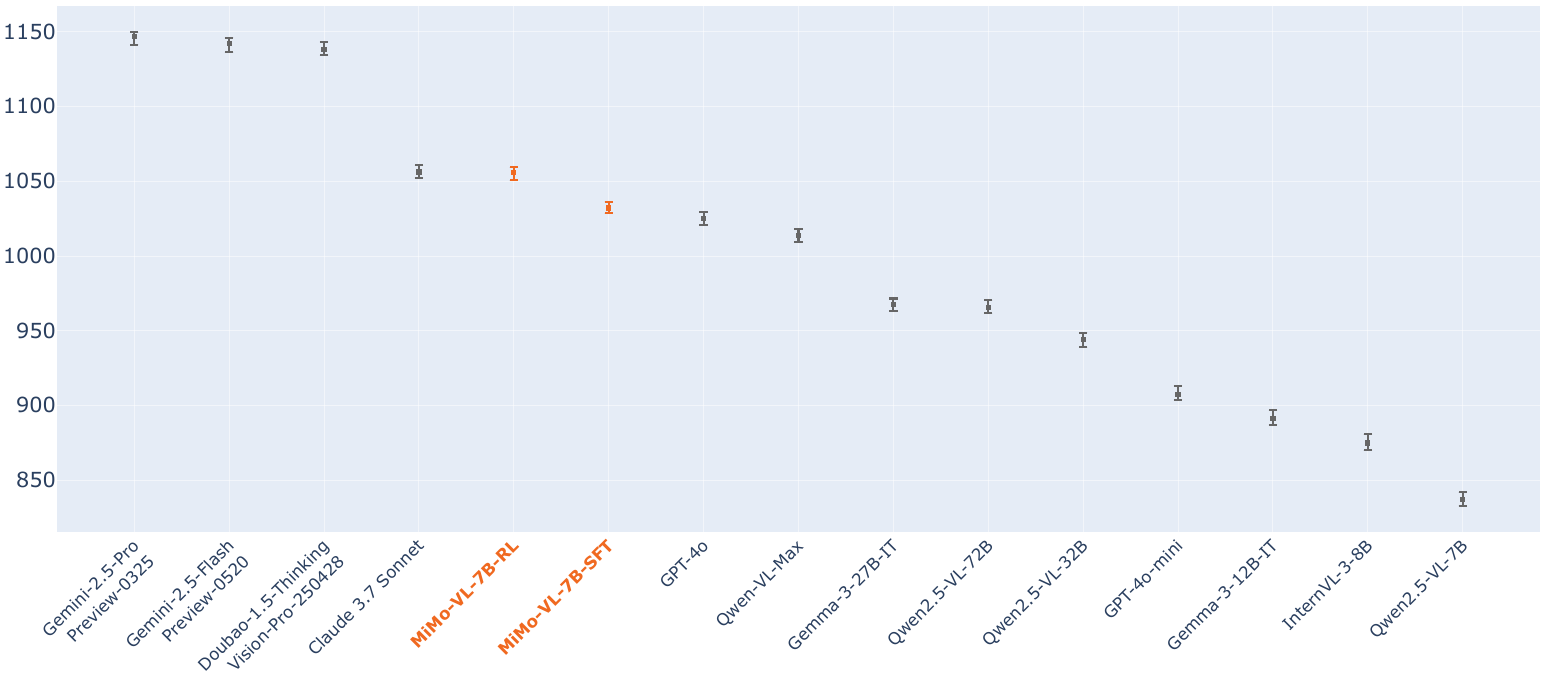}
    \caption{Elo ratings comparison across VLMs. \mimovlrl{} achieves the highest rating among open-source models, approaching the performance of proprietary alternatives such as Claude 3.7 Sonnet.}
    \label{fig:elo_rating}
\end{figure}

\begin{figure}
    \centering
    \includegraphics[width=0.95\linewidth]{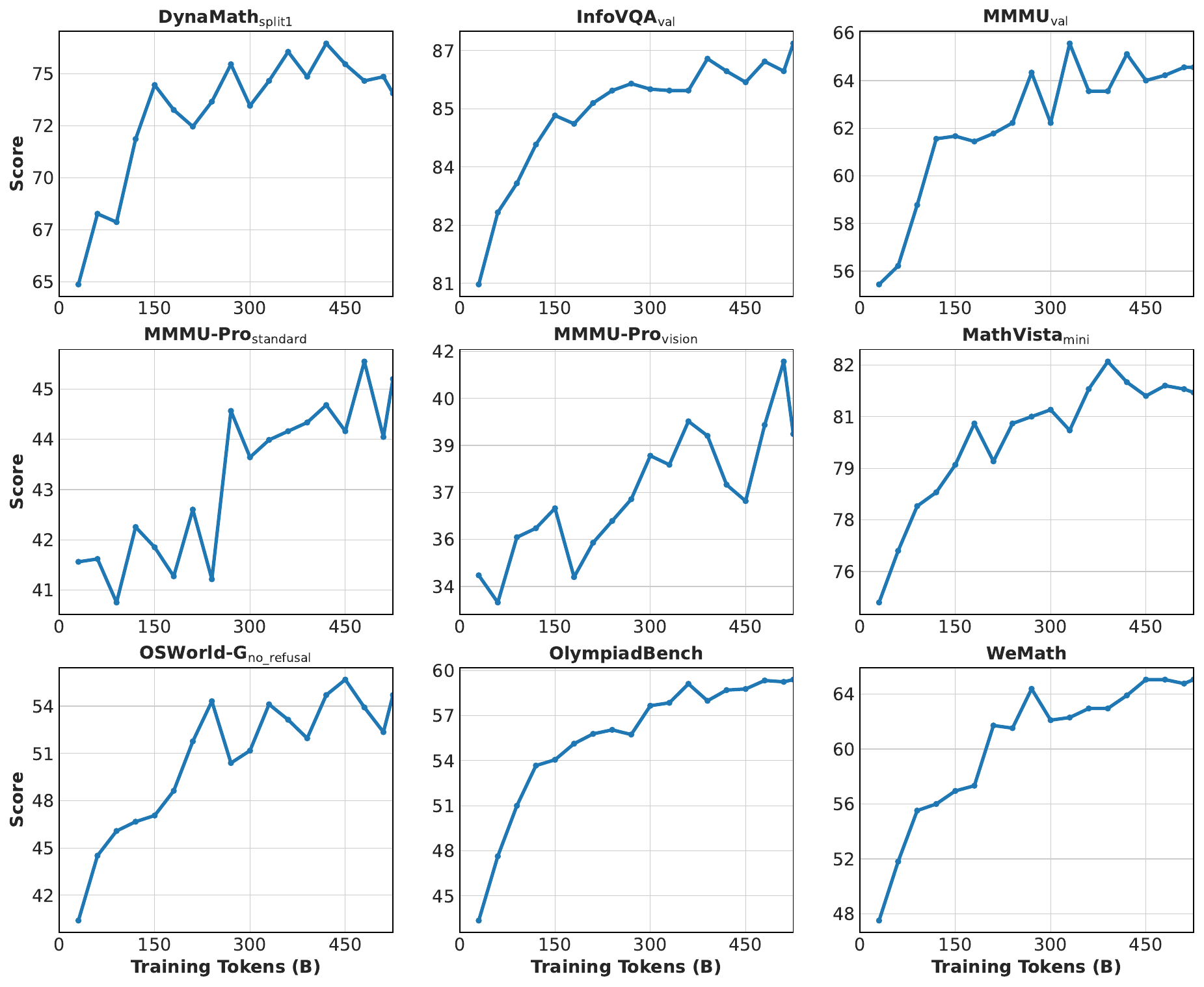}
    \caption{\mimovlsft{} training curves in Stage 4.}
    \label{fig:mimo_7b_sft_performance}
\end{figure}

\begin{figure}[t!]
    \centering
    \includegraphics[width=0.6\linewidth]{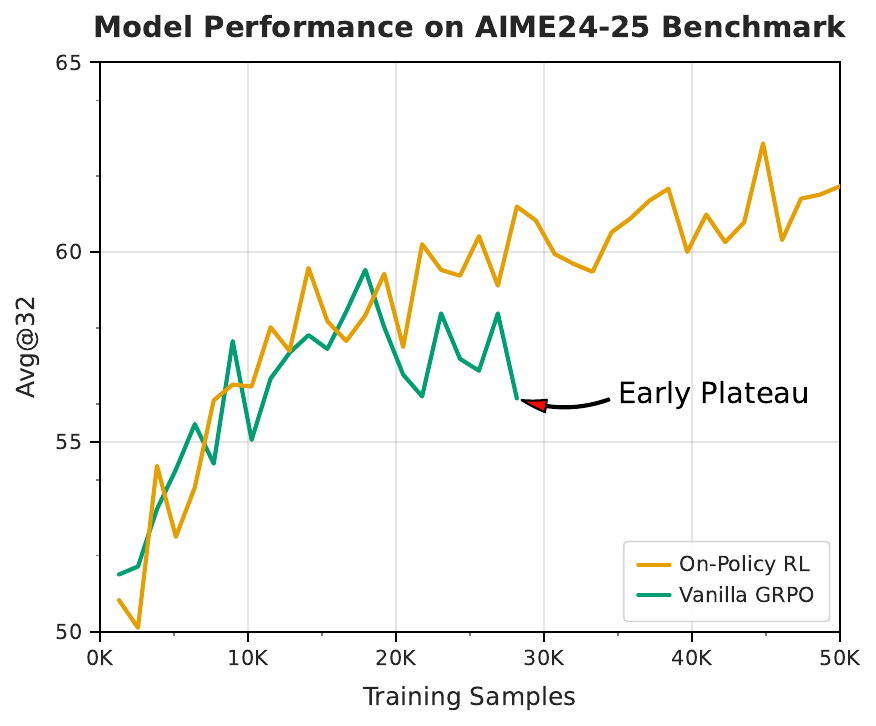}
    \caption{On-policy RL and vanilla GRPO shows contrasting scaling behavior: on-policy RL performance continuously improves with more data, while vanilla GRPO reaches a plateau around 20,000 samples.}
    \label{fig:onpolicyrl}
\end{figure}

\subsection{Elo Rating}
Inspired by ChatbotArena~\citep{zheng2023chatbotarena}, we construct a balanced bilingual (Chinese and English) in-house evaluation dataset comprising real user prompts. This approach assesses user preference beyond traditional benchmark scores, providing insights into practical model performance in real-world scenarios.

Following the methodology of \citep{Chou2024VisionArena2R}, we conduct pairwise comparisons between \mimovl{} and competing models, including leading proprietary models and open-source VLMs ranging from 7B to 72B parameters. We compute Elo ratings based on GPT-4o judgments with style-controlled evaluation protocols. 
Our evaluation covers a diverse range of visual-linguistic tasks, including multimodal reasoning, image understanding, and GUI interaction scenarios, therefore serving as a good proxy of user preference.

As illustrated in Figure~\ref{fig:elo_rating}, \mimovlrl{} achieves the highest Elo rating among all evaluated open-source VLMs, ranking first across models spanning from 7B to 72B parameters. This demonstrates superior user experience across the evaluation set, with our model's performance closely approaching that of proprietary models such as Claude 3.7 Sonnet. 
Moreover, MORL brings a boost of 22+ points for the \mimovl{}-SFT.
These results highlight the competitive capability of our models and validate the effectiveness of our training methodology.

\section{Discussion}

\subsection{Boosting Reasoning Capability in Pre-training}
Figure~\ref{fig:mimo_7b_sft_performance} shows the performance of \mimovlsft{} during Stage 4, its final pre-training phase. In this stage, substantial volumes of synthetic long-form reasoning data are incorporated, and model performance increases sharply, e.g., +9 on MMMU, +14 on OSWorld-G, and +16 on OlympiadBench. Notably, model performance continuously improves without saturation. 
These improvements are attributed to an increased depth in the model's reasoning. For instance, on  MMMU, the model's average number of response tokens grows from 680 to 2.5K per question after Stage 4, indicating a more detailed and profound level of reasoning when tackling problems.

\subsection{On-Policy RL v.s. Vanilla GRPO}

We explore the benefits of on-policy RL v.s. vanilla GRPO with text-only reasoning tasks.
As illustrated in Figure~\ref{fig:onpolicyrl}, the on-policy algorithm demonstrates a consistent positive correlation between training data volume and performance score. Its learning curve shows no signs of saturation within the observed training window, suggesting potential for further enhancement with additional computational resources and data. 
Conversely, the vanilla GRPO algorithm initially exhibits higher sample efficiency, achieving robust performance early in training. This advantage, however, is transient. The algorithm's performance generally saturates around 20,000 training samples, beyond which further training yields negligible improvements.

\begin{figure}[!htbp]
  \centering
  \begin{tabular}{m{15cm}}
  \toprule
  \begin{center}
  \includegraphics[width=0.55\linewidth]{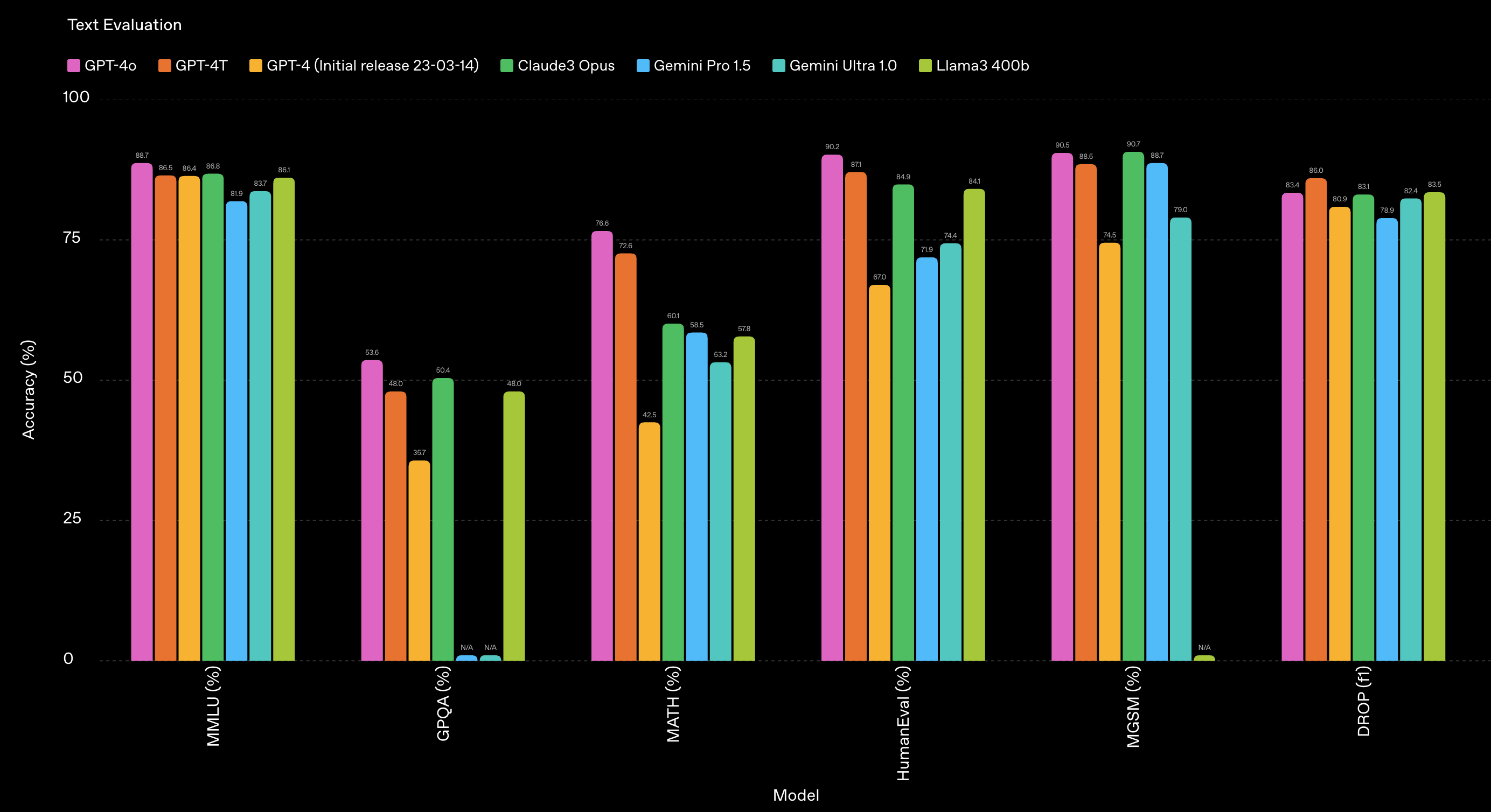}
  \end{center}\\
  {\footnotesize\textbf{Instruction:} Convert it to a table.}\\
  {\footnotesize\textbf{\mimovlrl{}:}} \\
  \begin{center}
  \includegraphics[width=0.75\linewidth]{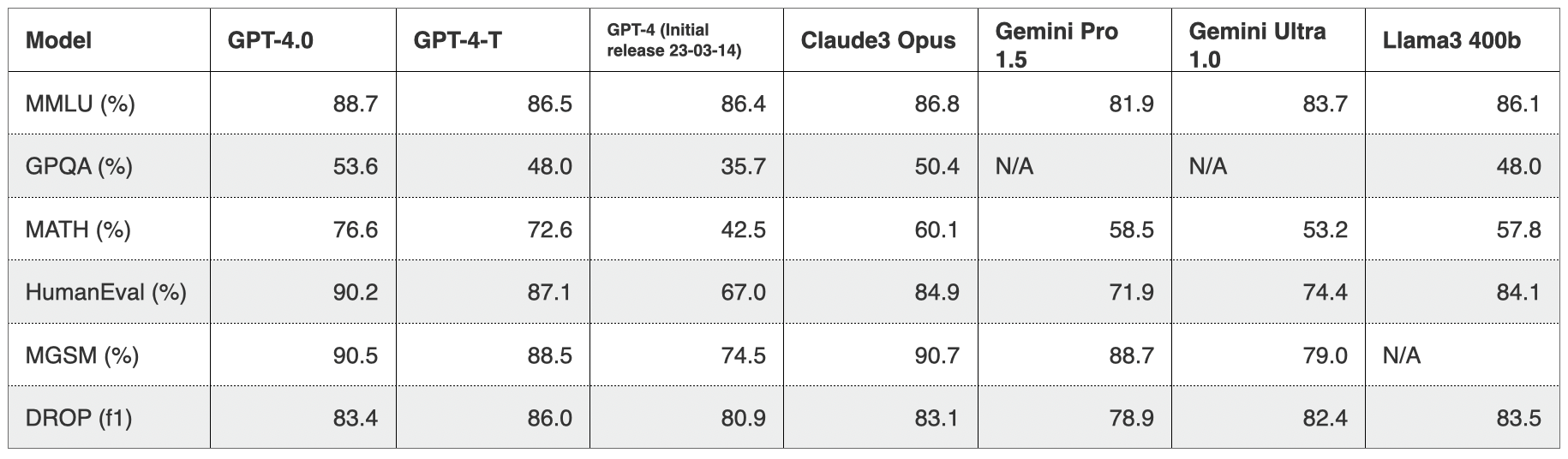}
  \end{center} \\
  
  \midrule
  
  \begin{center}
  \includegraphics[width=0.55\linewidth]{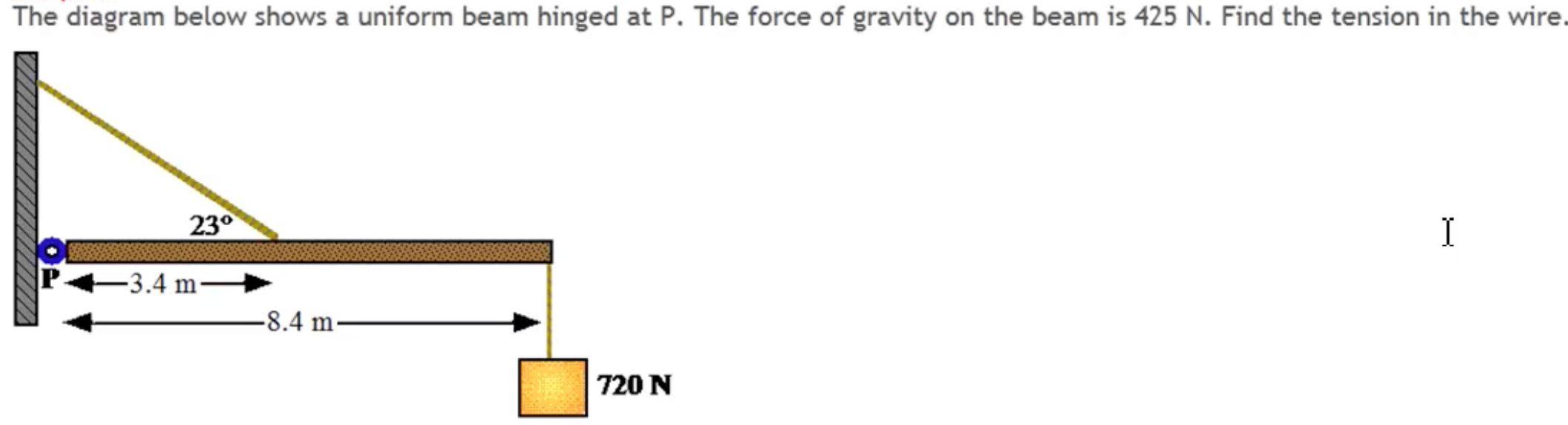}
  \end{center} \\
  {\footnotesize\textbf{Instruction:} Answer the question in image.} \\
  {\footnotesize \textbf{\mimovlrl{}:} \texttt{<think>...</think>} To solve for the tension in the wire, we use the \textbf{equilibrium of torques} about the hinge point $ P $. For a beam in equilibrium, the sum of clockwise torques equals the sum of counterclockwise torques.  

\textbf{Step 1: Identify Forces and Torques}

\quad - Beam’s weight ($ 425 \, \text{N} $): Acts downward at the beam’s midpoint (since it is uniform). The beam’s length is $ 8.4 \, \text{m} $, so the midpoint is $ \frac{8.4}{2} = 4.2 \, \text{m} $ from $ P $.  

\quad - Block’s weight ($ 720 \, \text{N} $): Acts downward at $ 8.4 \, \text{m} $ from $ P $. 

\quad - Tension in the wire ($ T $): Acts at $ 3.4 \, \text{m} $ from $ P $ at an angle of $ 23^\circ $ above the horizontal. Only the \textbf{vertical component} of tension ($ T \sin 23^\circ $) contributes to torque (the horizontal component acts along the beam, so its torque about $ P $ is zero).

\textbf{Step 2: Calculate Torques}

Torque is defined as $ \text{force} \times \text{perpendicular distance from pivot} $.  

\quad - Clockwise torques (from the beam and block weights):  
  - Beam’s weight: $ 425 \, \text{N} \times 4.2 \, \text{m} = 1785 \, \text{N·m} $  
  - Block’s weight: $ 720 \, \text{N} \times 8.4 \, \text{m} = 6048 \, \text{N·m} $  
  - Total clockwise torque: $ 1785 + 6048 = 7833 \, \text{N·m} $  

\quad - Counterclockwise torque (from tension’s vertical component):  
  - Tension’s vertical component: $ T \sin 23^\circ $  
  - Perpendicular distance from $ P $ to the wire: $ 3.4 \, \text{m} $  
  - Torque from tension: $ T \sin 23^\circ \times 3.4 \, \text{m} $

\textbf{Step 3: Solve for Tension} 

In equilibrium, clockwise torque = counterclockwise torque:  
$
T \sin 23^\circ \times 3.4 = 7833
$  

Solve for $ T $:  
$
T = \frac{7833}{3.4 \times \sin 23^\circ}
$  

Using $ \sin 23^\circ \approx 0.3907 $:  
$
T = \frac{7833}{3.4 \times 0.3907} \approx \frac{7833}{1.328} \approx 5900 \, \text{N}
$

\textbf{Final Answer}  
The tension in the wire is approximately $ \boldsymbol{5900 \, \text{N}} $.
}\\
  \bottomrule
  \end{tabular}\\
\caption{Examples of \mimovl{} solving complicated problems.}
\label{fig:case_main}
\end{figure}

\subsection{Interference Between RL Tasks}
While MORL training enhances performance on nearly all evaluated tasks, achieving stable and simultaneous improvements across diverse task domains remains a significant challenge.
During training, we observe that reasoning tasks exhibit disparities with visual perception and grounding tasks, such as visual grounding and counting, making it difficult to match the performance of standalone RL on individual tasks.

The potential cause lies in the opposing growth trends of response length: reasoning tasks encourage longer CoT during the RL process, whereas grounding and counting tasks lead to shrinking ones.
Disparities in task difficulty and the risk of reward hacking may also contribute to this interference.
We are actively investigating the underlying causes of this phenomenon and seeking solutions to achieve consistent and persistent growth across all tasks.

\section{Case Study}

\begin{figure}[!htbp]
    \centering
    \includegraphics[width=0.95\linewidth]{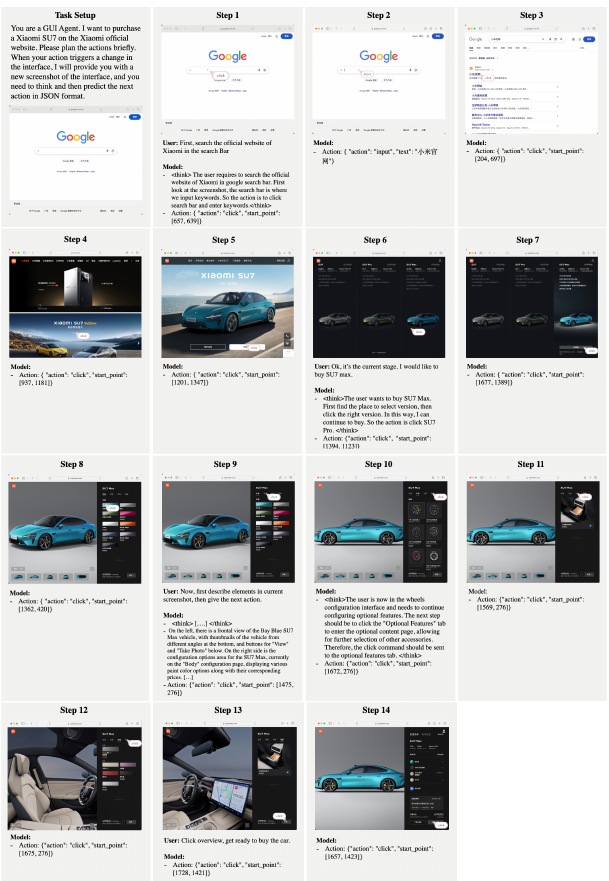}
    \caption{A case demonstrating the agentic capabilities of our model. \mimovl{} successfully navigates a website to add the Xiaomi SU7 to the wishlist, customizing both paint and interior options. All screenshots are of size 1886*1544 (width*height).}
    \label{fig:gui_case}
\end{figure}

We present qualitative results in Figure~\ref{fig:case_main} and Appendix~\ref{sec:more_cases}.
As depicted in the top example in Figure~\ref{fig:case_main}, our model showcases strong plot understanding capabilities, successfully converting an intricate plot into a well-structured markdown table.
Additionally, we highlight the models' superior reasoning capabilities in STEM tasks.
In the examples shown in Figure~\ref{fig:case1} and Figure~\ref{fig:case4}, the model effectively addresses multiple STEM questions within a single response.
Furthermore, our model exhibits strong agentic capabilities.
As illustrated in Figure~\ref{fig:gui_case}, \mimovl{} successfully navigates a website to add the Xiaomi SU7 to the wishlist, with customized paint and interior options.

\clearpage
\section{Conclusions}
In this report, we present our efforts in building \mimovl{} models.
Leveraging curated high-quality pre-training datasets and our MORL framework, \mimovlsft{} and \mimovlrl{} demonstrate state-of-the-art performance across evaluated benchmarks. 
We share key observations from our development process: the consistent performance gains from incorporating reasoning data in later pre-training stages, the advantages of on-policy RL over vanilla GRPO, and the challenges of task interference when applying MORL across diverse capabilities.
Alongside the released model checkpoints, we open-source our comprehensive evaluation suite to promote transparency and reproducibility in multimodal research. 
We hope our work advances the development of capable open-source vision-language models and provides valuable insights for the community.

\bibliography{main}

\appendix
\newpage

\section{Contributions and Acknowledgments}
We would like to express our sincere gratitude to all contributors for their invaluable support and efforts, including the Xiaomi LLM-Plus, Mify, MiChat and CloudML teams, as well as those not explicitly listed in this paper.
Authors within each role are listed in \textit{\textbf{reverse order}} by their first names. 

\definecolor{ourblue}{RGB}{0, 0, 0}
\definecolor{ourgreen}{RGB}{0, 0, 0}
\definecolor{ourred}{RGB}{0, 0, 0}

\begin{multicols}{2} %
\noindent
\textbf{\color{ourred} Core Contributors} \\
\color{ourred} Zihao Yue \\
\color{ourred} Zhenru Lin \\
\color{ourred} Yifan Song \\
\color{ourred} Weikun Wang \\
\color{ourred} Shuhuai Ren \\
\color{ourred} Shuhao Gu \\
\color{ourred} Shicheng Li \\
\color{ourred} Peidian Li \\
\color{ourred} Liang Zhao \\
\color{ourred} Lei Li \\
\color{ourred} Kainan Bao \\
\color{ourred} Hao Tian \\
\color{ourred} Hailin Zhang \\
\color{ourred} Gang Wang \\
\color{ourred} Dawei Zhu \\
\color{ourred} Cici \\
\color{ourred} Chenhong He \\
\color{ourred} Bowen Ye \\
\color{ourred} Bowen Shen \\

\noindent
\textbf{\color{ourblue} Contributors} \\
\color{ourblue} Zihan Zhang \\
\color{ourblue} Zihan Jiang \\
\color{ourblue} Zhixian Zheng \\
\color{ourblue} Zhichao Song \\
\color{ourblue} Zhenbo Luo \\
\color{ourblue} Yue Yu \\
\color{ourblue} Yudong Wang \\
\color{ourblue} Yuanyuan Tian \\
\color{ourblue} Yu Tu \\
\color{ourblue} Yihan Yan \\
\color{ourblue} Yi Huang \\
\color{ourblue} Xu Wang \\
\color{ourblue} Xinzhe Xu \\
\color{ourblue} Xingchen Song \\
\color{ourblue} Xing Zhang \\
\color{ourblue} Xing Yong \\
\color{ourblue} Xin Zhang \\
\color{ourblue} Xiangwei Deng \\
\color{ourblue} Wenyu Yang \\
\color{ourblue} Wenhan Ma \\
\color{ourblue} Weiwei Lv \\
\color{ourblue} Weiji Zhuang \\
\color{ourblue} Wei Liu  \\
\color{ourblue} Sirui Deng \\
\color{ourblue} Shuo Liu \\
\color{ourblue} Shimao Chen \\
\color{ourblue} Shihua Yu \\
\color{ourblue} Shaohui Liu \\
\color{ourblue} Shande Wang  \\
\color{ourblue} Rui Ma \\
\color{ourblue} Qiantong Wang \\
\color{ourblue} Peng Wang \\
\color{ourblue} Nuo Chen \\
\color{ourblue} Menghang Zhu \\
\color{ourblue} Kangyang Zhou \\
\color{ourblue} Kang Zhou  \\
\color{ourblue} Kai Fang \\
\color{ourblue} Jun Shi \\
\color{ourblue} Jinhao Dong \\
\color{ourblue} Jiebao Xiao \\
\color{ourblue} Jiaming Xu  \\
\color{ourblue} Huaqiu Liu \\
\color{ourblue} Hongshen Xu \\
\color{ourblue} Heng Qu \\
\color{ourblue} Haochen Zhao \\
\color{ourblue} Hanglong Lv \\
\color{ourblue} Guoan Wang \\
\color{ourblue} Duo Zhang \\
\color{ourblue} Dong Zhang \\
\color{ourblue} Di Zhang \\
\color{ourblue} Chong Ma \\
\color{ourblue} Chang Liu \\
\color{ourblue} Can Cai \\
\color{ourblue} Bingquan Xia \\

\end{multicols} %

\section{Model Configuration of MiMo-VL-7B}
\label{sec:model_config}
In Table~\ref{tab:model_config}, the architecture and configuration of \mimovl{} are detailed.
\begin{table}[h]
\centering
\begin{tabular}{lcc}
\toprule
                  & Vision Encoder & Language Model             \\ \midrule
\# Layers         & 32             & 36                         \\
\# Heads          & 16             & 32                         \\
Hidden Size       & 1280           & 4096                       \\
Intermediate Size & 3456           & 11008                      \\ 
Position Embedding & 2D RoPE       & MRoPE~\citep{bai2025qwen2} \\
Patch Size        & 14             & -                          \\
\bottomrule
\end{tabular}
\caption{Configuration of \mimovl{}. We adopt Qwen2.5-ViT~\citep{Qwen25VL} as our visual encoder to support native resolution inputs, and \mimobase{}~\citep{xia2025mimo} as our LLM backbone to leverage its strong reasoning capability. Compared to the LLM backbone of Qwen2.5-VL-7B~\citep{Qwen25VL}, our LLM differs in its number of layers (36 vs. 28), hidden size (4096 vs. 3584), and intermediate size (11008 vs. 18944).}
\label{tab:model_config}

\end{table}

\section{Evaluation Benchmarks}
\label{sec:eval_bench}

We evaluate our models across 50 diverse tasks, including:

\paragraph{General Visual Understanding}
AI2D~\citep{kembhavi2016diagram}, 
BLINK~\citep{fu2024blink}.
CV-Bench~\citep{tong2024cambrian}, 
MMMU~\citep{yue2024mmmu}, 
MMMU-Pro (Standard and Vision)~\citep{yue2024mmmupro}, 
Mantis~\citep{jiang2024mantis},
MME-RealWorld (English and Chinese)~\citep{mmerealworld},
MMBench (English and Chinese)~\citep{liu2024mmbench}, 
VibeEval~\citep{padlewski2024vibe},
VL-RewardBench~\citep{Li2024VLRewardBenchAC},
V\textsuperscript{*}~\citep{wu2023vguidedvisualsearch}, 
and VLMs are Blind~\citep{rahmanzadehgervi2025visionlanguagemodelsblind}.

\paragraph{General Grounding and Counting}
RefCOCO~\citep{refcoco,refcocog,referit}, 
CountBench~\citep{paiss2023teaching}, and
PixmoCount~\citep{deitke2024molmo}

\paragraph{Document and Chart Understanding} 
ChartQA~\citep{masry2022chartqa}, InfographicVQA~\citep{mathew2022infographicvqa}, DocVQA~\citep{mathew2021docvqa}, 
OCRBench~\citep{liu2024ocrbench},
SEED-Bench-2-Plus~\citep{seedbench2plus},
and CharXiv (RQ/DQ)~\citep{wang2024charxiv}.

\paragraph{Video Understanding and Localization}
Video-MME~\citep{fu2024video}, 
Video-MMMU~\citep{hu2025video},
EgoSchema~\citep{mangalam2023egoschema},
and Charades-STA~\cite{charadessta}.

\paragraph{GUI Understanding and Grounding}
WebSrc~\citep{chen2021websrc}, VisualWebBench~\citep{liu2024visualwebbench}, ScreenSpot~\citep{cheng2024seeclick}, 
ScreenSpot-V2~\citep{wu2024atlas}, 
ScreenSpot-Pro~\citep{li2025screenspot}, 
and OSWorld-G~\citep{osworldg}.

\paragraph{Text-only Benchmarks}
GPQA~\citep{rein2024gpqa}, SuperGPQA~\citep{du2025supergpqa}, DROP~\citep{dua2019drop}, MMLU-Pro~\citep{wang2024mmlu}, 
and IFEval~\citep{zhou2023instructionfollowingevaluationlargelanguage}.

\paragraph{Multimodal Reasoning} 
OlympiadBench~\citep{he2024olympiadbench}, MathVision~\citep{wang2024measuring}, MathVerse (Vision Only)~\citep{mathverse}, DynaMath~\citep{zou2024dynamath}, WeMath~\citep{qiao2024we}, LogicVista~\citep{xiao2024logicvista}, 
and MathVista~\citep{lu2023mathvista}. 

\paragraph{Text Reasoning}
MATH500~\citep{hendrycks2020measuring}, AIME 2024~\citep{AIME}, and AIME 2025~\citep{AIME25}.

\section{GUI Action Space}
\label{sec:gui_action_space}
The syntax and definition for each action in the GUI action space are summarized in Table ~\ref{tab:action_details_custom}.

\begin{table}[h!]
\centering
\setlength{\extrarowheight}{3pt} %
\begin{tabular}{@{}ll@{}} 
\hline %
click & \textit{Syntax:} \texttt{\{"action":"click","start\_point":[x,y],"text"(opt.):text\}} \\
      & \textit{Definition:} Click at (x, y) on an element with the given text. \\
\hline %
scroll & \textit{Syntax:} 
\texttt{\{"action":"scroll","direction":dir,"scroll\_distance"(opt.):dist\}}
\\
       & \textit{Definition:} Scroll in the given direction by a specified distance. \\
       & \textit{Notes:} Scroll up/down → to see more below/above. The same logic applies horizontally. \\
\hline %
input & \textit{Syntax:} 
\texttt{\{"action":"input","text":text,"start\_point"(opt.):[x,y]\}} \\
      & \textit{Definition:} Type the specified text at coordinates (x, y). \\
\hline %
drag & \textit{Syntax:} 
\texttt{\{"action":"drag","start\_point":[x1,y1],"end\_point":[x2,y2]\}} \\
     & \textit{Definition:} Drag from the \texttt{start\_point} to the \texttt{end\_point}.\\
\hline %
open & \textit{Syntax:} 
\texttt{\{"action":"open","app":app\_name\}} \\
     & \textit{Definition:} Open the specified application \texttt{app\_name}.\\
\hline %

press & \textit{Syntax:} 
\texttt{\{"action":"press","keys":[key1,key2,...]\}} \\
      & \textit{Definition:} Press the specified hotkeys (\texttt{[key1, key2,...]}).\\
\hline %

finished & \textit{Syntax:} 
\texttt{\{"action":"finished","status":status\}} \\
         & \textit{Definition:} Mark the task as complete with a given \texttt{status}.\\
\hline %
longpress & \textit{Syntax:} 
\texttt{\{"action":"longpress","start\_point":[x,y]\}} \\
          & \textit{Definition:} Long press at coordinates (x,y). \\
\hline %

hover & 
\textit{Syntax:} 
\texttt{\{"action":"hover"\}} \\
      & \textit{Definition:} Hover the mouse over a location. \\
\hline %
select & 
\textit{Syntax:} 
\texttt{\{"action":"select","text":text\}} \\
       & \textit{Definition:} Select the specified \texttt{text}. \\
\hline %
wait & 
\textit{Syntax:}
\texttt{\{"action":"wait"\}} \\
     & \textit{Definition:} Pause for a brief moment. \\
\hline %
appswitch & 
\textit{Syntax:} 
\texttt{\{"action":"appswitch","app":app\_name\}} \\
          & \textit{Definition:} Switch to the specified application \texttt{app\_name}.\\
\hline %
\end{tabular}
\caption{Action Space Details. \textit{opt.} denotes \textit{optional}.}
\label{tab:action_details_custom}
\end{table}

\section{More Qualitative Examples}
\label{sec:more_cases}
\clearpage
\renewcommand{\arraystretch}{1.5}
\begin{CJK*}{UTF8}{gbsn}
\begin{figure}[h]
  \centering
  \begin{tabular}{m{15cm}}
  \toprule
  \textbf{Input image:}
  \begin{center}
  \includegraphics[width=7cm]{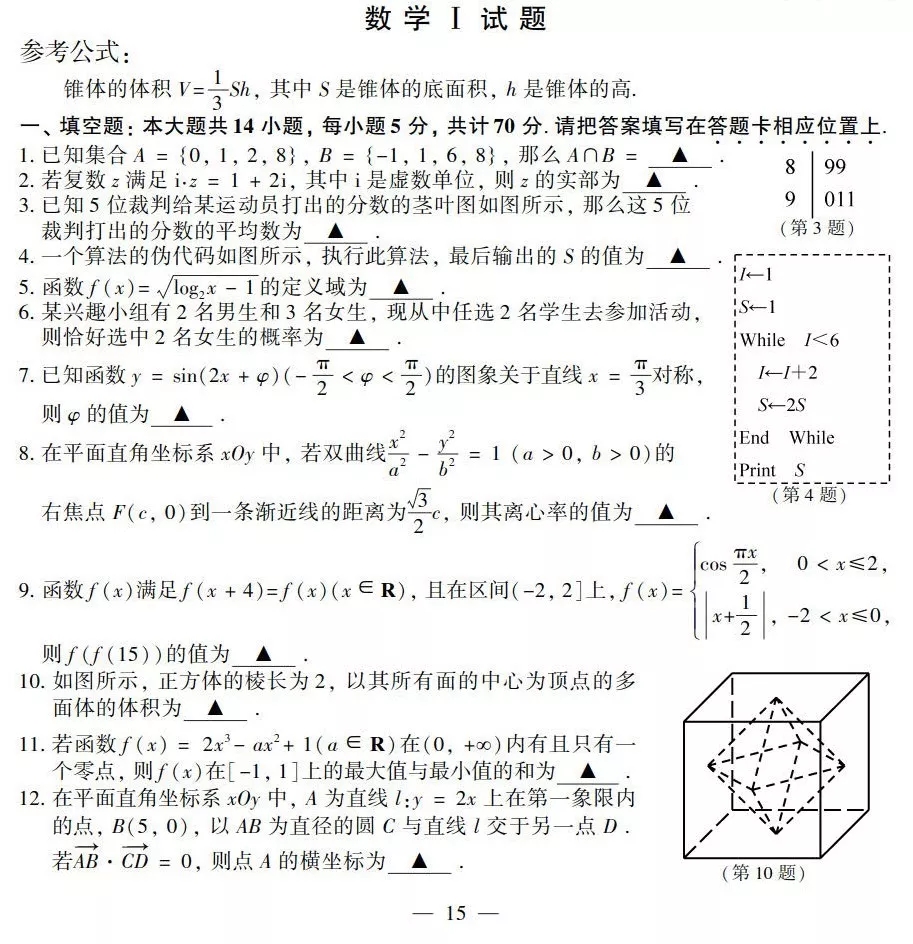}
  \end{center}
  \\
  {\footnotesize \textbf{Instruction:} 完成该试卷\newline\newline
  \textbf{\mimovl{}:} \texttt{<think>...</think>}
  
    \textbf{第1题：集合的交集}
    集合 $ A = \{0, 1, 2, 8\} $，$ B = \{-1, 1, 6, 8\} $，交集为同时属于两个集合的元素，即 $ A \cap B = \{1, 8\} $。

    \textbf{第2题：复数的实部} 
    设 $ z = x + yi $（$ x, y $ 为实数），由 $ i \cdot z = 1 + 2i $，即 $ i(x + yi) = -y + xi = 1 + 2i $。  
    分离实部和虚部：$ -y = 1 $（实部），$ x = 2 $（虚部）。故 $ z $ 的实部为 $ 2 $。

    \textbf{第3题：茎叶图的平均数}  
    茎叶图中，茎为十位，叶为个位。分数为 $ 89, 89, 90, 91, 91 $（共5个）。  
    平均数为 $ \frac{89 + 89 + 90 + 91 + 91}{5} = \frac{450}{5} = 90 $。

    \textbf{第4题：伪代码的循环结果}
    初始化 $ I = 1 $，$ S = 1 $。循环条件 $ I < 6 $：  
    - 第1次循环：$ I = 1 + 2 = 3 $，$ S = 2 \times 1 = 2 $；  
    - 第2次循环：$ I = 3 + 2 = 5 $，$ S = 2 \times 2 = 4 $；  
    - 第3次循环：$ I = 5 + 2 = 7 $（不满足 $ I < 6 $，退出循环）。  
    最终 $ S = 8 $。

    \textbf{第5题：函数的定义域}
    函数 $ f(x) = \sqrt{\log_2 x - 1} $ 需满足：  
    1. 二次根式内非负：$ \log_2 x - 1 \geq 0 $；  
    2. 对数有定义：$ x > 0 $。  
    由 $ \log_2 x \geq 1 $ 得 $ x \geq 2^1 = 2 $，故定义域为 $ [2, +\infty) $。

    \textbf{第6题：组合的概率}
    2男3女中选2人，总选法 $ C_5^2 = 10 $；选中2女生的选法 $ C_3^2 = 3 $。  
    概率为 $ \frac{3}{10} $。

    \textbf{第7题：正弦函数的对称轴}
    函数 $ y = \sin(2x + \varphi) $ 关于 $ x = \frac{\pi}{3} $ 对称，即 $ 2 \cdot \frac{\pi}{3} + \varphi = k\pi + \frac{\pi}{2} $（$ k \in \mathbb{Z} $）。  
    取 $ k = 0 $，得 $ \varphi = -\frac{\pi}{6} $（满足 $ -\frac{\pi}{2} < \varphi < \frac{\pi}{2} $）。

    \textbf{第8题：双曲线的离心率}
    双曲线 $ \frac{x^2}{a^2} - \frac{y^2}{b^2} = 1 $，右焦点 $ F(c, 0) $（$ c = \sqrt{a^2 + b^2} $），渐近线 $ bx - ay = 0 $。  
    焦点到渐近线的距离为 $ \frac{|bc|}{\sqrt{a^2 + b^2}} = b $，由题设 $ b = \frac{\sqrt{3}}{2}c $。  
    结合 $ c^2 = a^2 + b^2 $，得 $ a = \frac{c}{2} $，离心率 $ e = \frac{c}{a} = 2 $。

    \textbf{第9题：周期函数与复合函数}
    $ f(x + 4) = f(x) $ 说明周期为4。  
    - $ f(15) = f(15 - 4 \times 3) = f(3) = f(-1) $（$ 3 = -1 + 4 $）；  
    - $ x \in (-2, 0] $ 时，$ f(-1) = |-1 + \frac{1}{2}| = \frac{1}{2} $；  
    - $ f(\frac{1}{2}) = \cos\left( \frac{\pi}{2} \cdot \frac{1}{2} \right) = \cos\frac{\pi}{4} = \frac{\sqrt{2}}{2} $（$ \frac{1}{2} \in (0, 2] $）。

    \textbf{第10题：多面体的体积}
    正方体棱长2，面中心坐标为 $ (\pm 1, 0, 0), (0, \pm 1, 0), (0, 0, \pm 1) $，构成正八面体。  
    正八面体体积为两个四棱锥之和，底面正方形边长 $ \sqrt{2} $，高1，体积为 $ 2 \times \frac{1}{3} \times 2 \times 1 = \frac{4}{3} $。

    \textbf{第11题：函数的极值与零点}
    函数 $ f(x) = 2x^3 - ax^2 + 1 $ 在 $ (0, +\infty) $ 内有且只有一个零点，导数 $ f'(x) = 6x^2 - 2ax $，临界点 $ x = 0 $ 和 $ x = \frac{a}{3} $。  
    当 $ f(\frac{a}{3}) = 0 $ 时（极小值为0，仅1个零点），解得 $ a = 3 $。  
    $ f(x) $ 在 $ [-1, 1] $ 上，$ f(0) = 1 $（最大值），$ f(-1) = -4 $（最小值），和为 $ 1 + (-4) = -3 $。

    \textbf{第12题：向量垂直的条件}
    设 $ A(x, 2x) $，$ AB = (5 - x, -2x) $，圆心 $ C\left( \frac{x + 5}{2}, x \right) $，另一交点 $ D(1, 2) $（由圆与直线交点唯一性得）。  
    向量 $ \overrightarrow{CD} = \left( \frac{-x - 3}{2}, 2 - x \right) $，由 $ \overrightarrow{AB} \cdot \overrightarrow{CD} = 0 $，解得 $ x = 3 $。

    \textbf{答案}
    
    1. $ \boldsymbol{\{1, 8\}} $  
    2. $ \boldsymbol{2} $  
    3. $ \boldsymbol{90} $  
    4. $ \boldsymbol{8} $  
    5. $ \boldsymbol{[2, +\infty)} $  
    6. $ \boldsymbol{\frac{3}{10}} $  
    7. $ \boldsymbol{-\frac{\pi}{6}} $  
    8. $ \boldsymbol{2} $  
    9. $ \boldsymbol{\frac{\sqrt{2}}{2}} $  
    10. $ \boldsymbol{\frac{4}{3}} $  
    11. $ \boldsymbol{-3} $  
    12. $ \boldsymbol{3} $
    }
  \\
  \bottomrule
  \end{tabular}
\caption{Example of \mimovlrl{} answering all questions within a whole test paper.}
\label{fig:case1}
\end{figure}
\renewcommand{\arraystretch}{1}
\end{CJK*}
\clearpage

\renewcommand{\arraystretch}{1.5}
\begin{CJK*}{UTF8}{gbsn}
\begin{figure}[h]
  \centering
  \begin{tabular}{m{7cm}m{8cm}}
  \toprule
  \includegraphics[width=7cm]{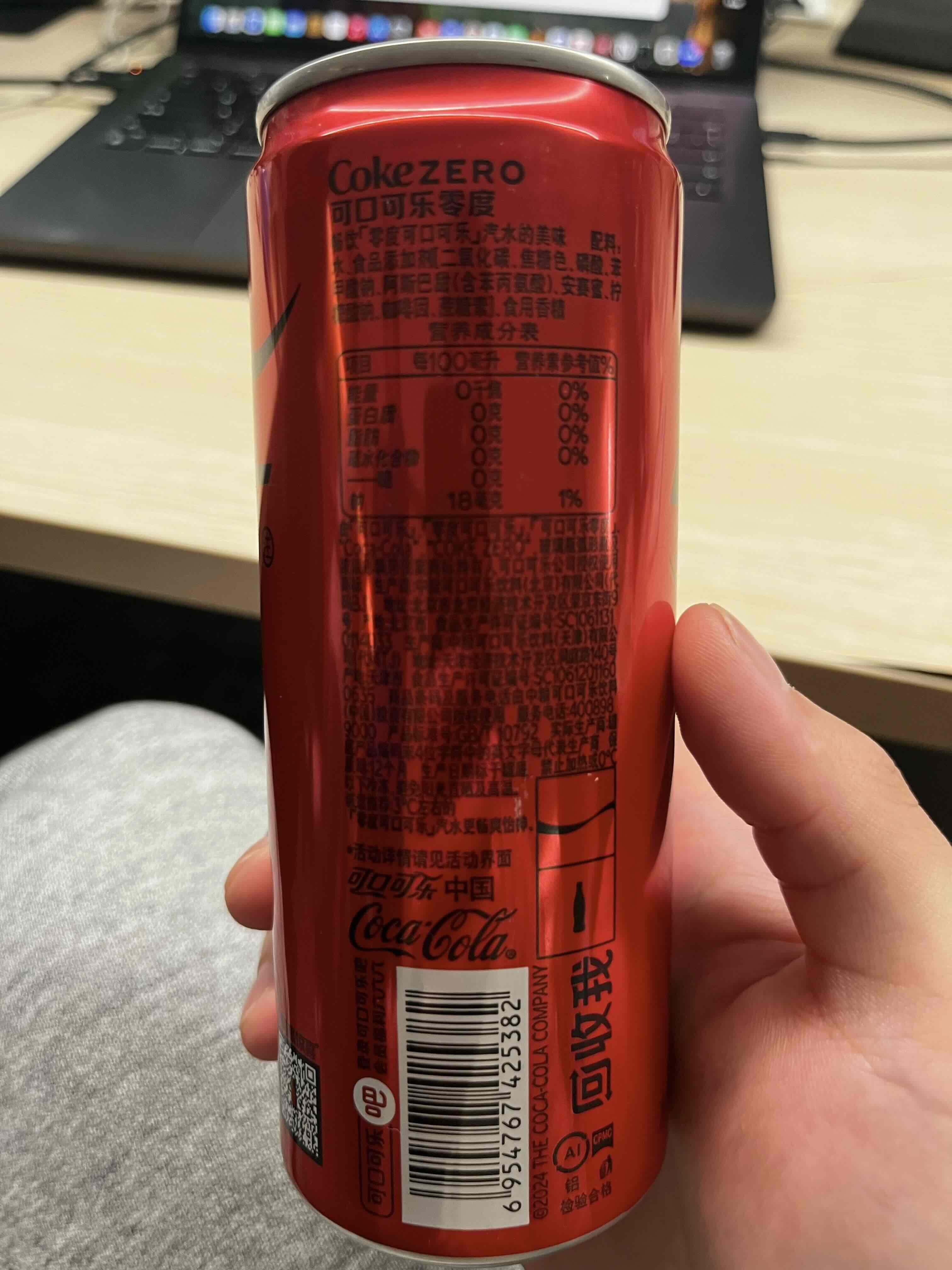} & 
  {\textbf{Instruction:} 这瓶可乐喝10罐会摄入多少克钠？\par
  \textbf{\mimovlrl{}:} \texttt{<think>...</think>}要计算喝10罐可口可乐零度时摄入的钠量，需按以下步骤分析：  
  \begin{enumerate}
      \item \textbf{步骤1：明确营养成分表信息}
    从图片中“营养成分表”可知：\textbf{每100毫升}该饮料中，钠的含量为 \textbf{18毫克}。  
      \item \textbf{步骤2：确定单罐饮料容量（行业常见规格）}
    可口可乐零度的罐装规格通常为 \textbf{330毫升/罐}（这是饮料行业的常见罐装容量，若实际容量不同需以产品标注为准，此处默认330ml）。
     \item \textbf{步骤3：计算单罐钠含量}
    单罐（330ml）中钠的含量 = 每100ml钠含量 × 单罐容量 ÷ 100  
    即：$ 18\ \text{毫克/100ml} × 330\ \text{ml} ÷ 100 = 59.4\ \text{毫克} $  
    \item \textbf{步骤4：计算10罐钠总量} 
    10罐钠总量 = 单罐钠含量 × 10  
    即：$ 59.4\ \text{毫克} × 10 = 594\ \text{毫克} $ 
  \end{enumerate}
  \par
  因此，喝10罐该款可口可乐零度，约摄入 \textbf{594毫克} 钠。  } \\
  \hline
  \includegraphics[width=7cm]{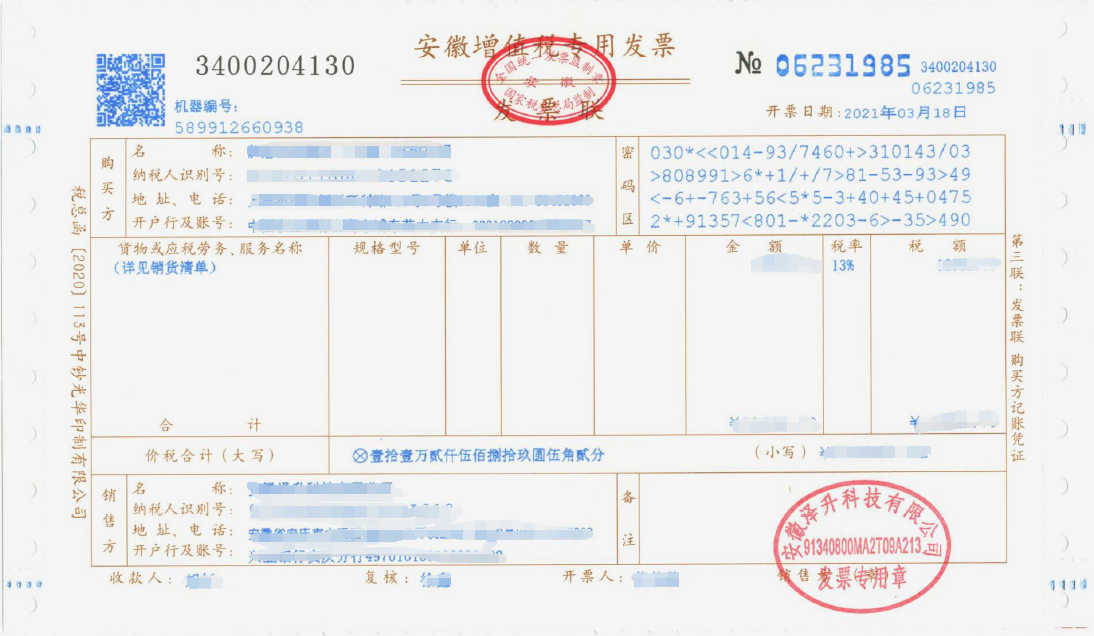} & 
  {\textbf{Instruction:} 该发票缺失的税额应该是多少？
\par
  \textbf{\mimovlrl{}:} \texttt{<think>...</think>}要计算该发票的税额，需结合\textbf{税率}和\textbf{价税合计金额}来推导： 
  \begin{enumerate}
      \item \textbf{步骤1：明确关键信息}
    - 发票税率：$13\%$（发票中“税率”栏显示）  
    - 价税合计金额（大写）：壹拾壹万贰仟伍佰捌拾玖圆伍角贰分，对应小写为 \textbf{112589.52元}
    \item \textbf{步骤2：税额的计算逻辑}  
    发票税额 = 价税合计金额 × （1 - 税率）× 税率   （即：总价价税合计 = 不含税金额 + 税额 → 税额 = 价税合计 × 税率 ÷ (1 + 税率) ）  
    \end{enumerate}
    设不含税金额为 $\ X $ ，则：  
    $ X + X×13\% = 112589.52 $
    $ X×1.13 = 112589.52 $  
    $ X = 112589.52 ÷ 1.13 = 99636.74 $ 元【不含税金额】  
    
    税额 = 不含税金额 × 税率 =  99636.74 × 13\% = 12952.78 元  
} \\
  \bottomrule
  \end{tabular}
\caption{Example of \mimovlrl{} solving complex OCR problems with reasoning.}
\label{fig:case2}
\end{figure}
\end{CJK*}
\renewcommand{\arraystretch}{1}

\clearpage

\renewcommand{\arraystretch}{1.5}
\begin{figure}[h]
  \centering
  \begin{tabular}{m{7cm}m{8cm}}
  \toprule
  \includegraphics[width=7cm]{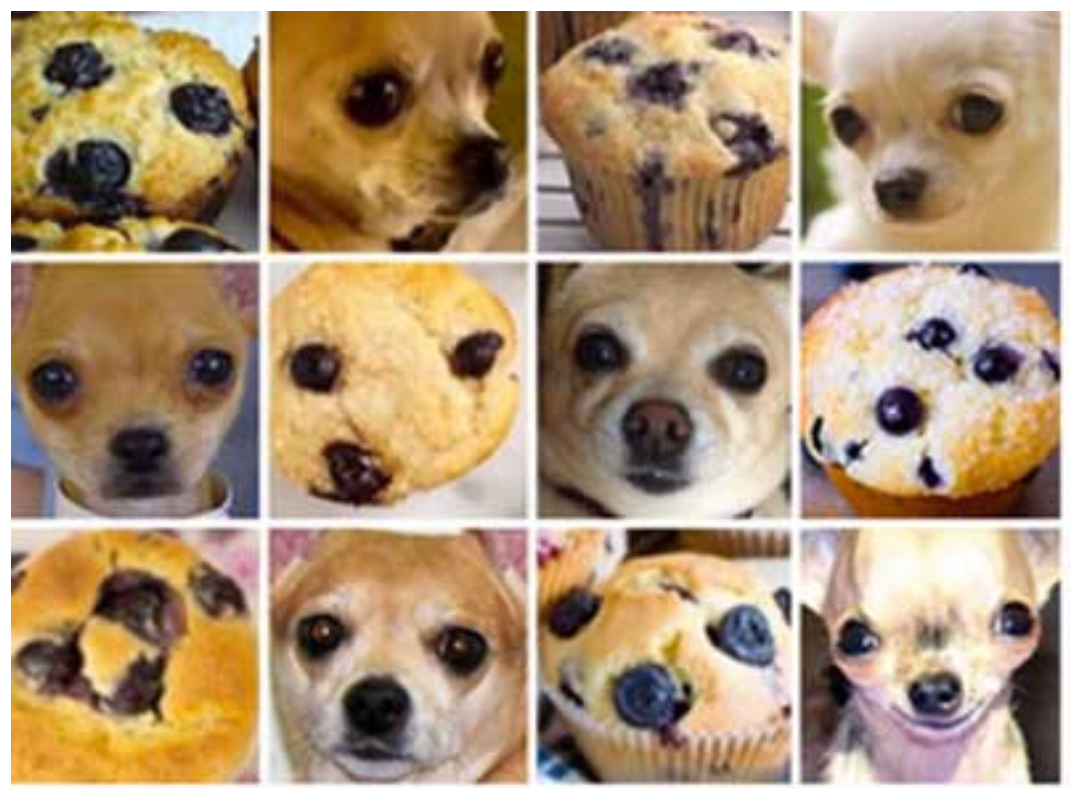} & 
  {\textbf{Instruction:} How many muffin in the image?\par
  \textbf{\mimovlrl{}:} \texttt{<think>...</think>}
To determine the number of muffins in the image, we analyze the 3×4 grid (12 cells total) and count muffins row by row:  

- \textbf{Top row (row 1)}: Columns 1 and 3 are muffins → 2 muffins.  
- \textbf{Middle row (row 2)}: Columns 2 and 4 are muffins → 2 muffins.
- \textbf{Bottom row (row 3)}: Columns 1 and 3 are muffins → 2 muffins. 

Adding these, \( 2 + 2 + 2 = 6 \).

There are 6 muffins in the image.} \\
\midrule
  \includegraphics[width=7cm]{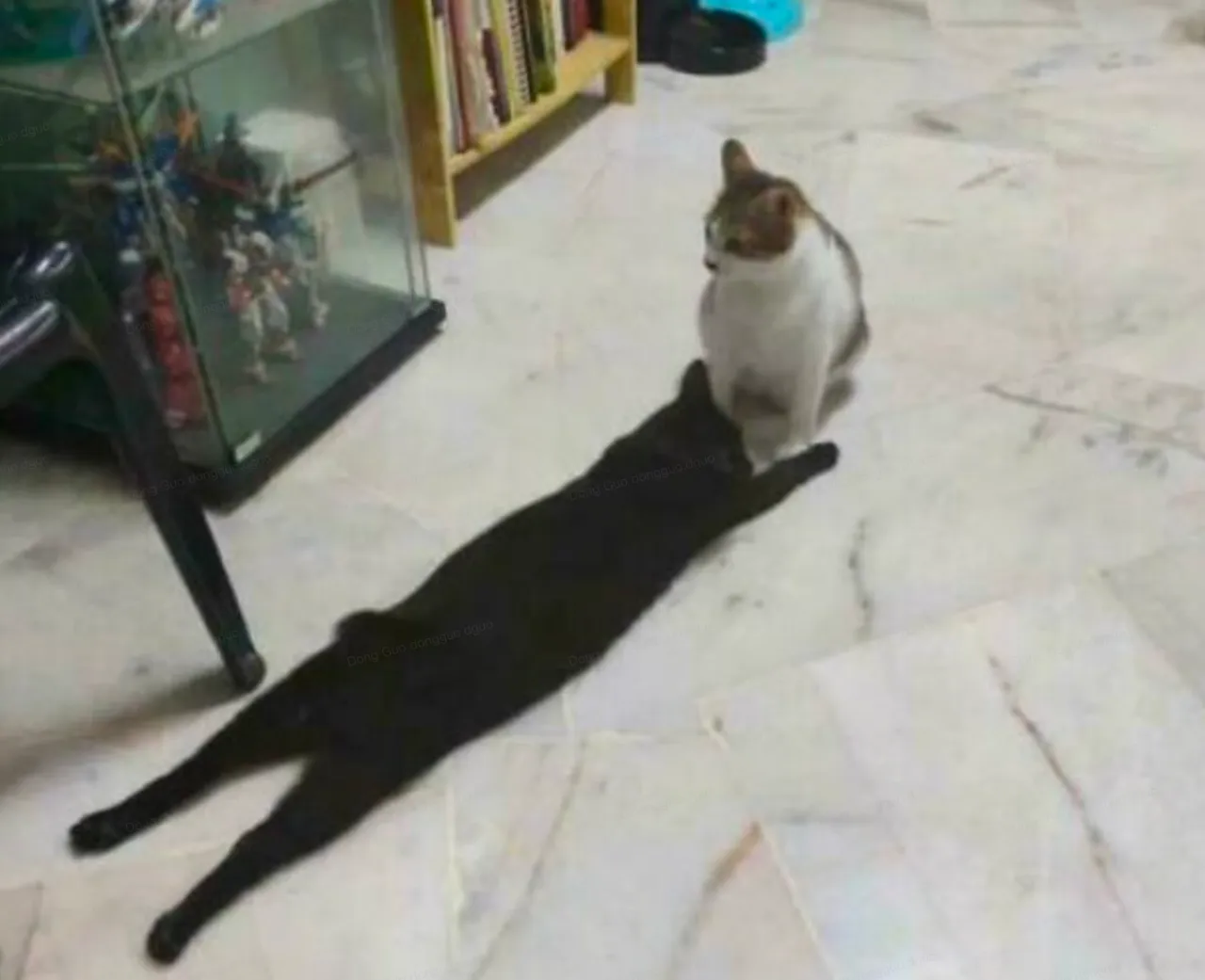} & 
  {\textbf{Instruction:} How many cats in the image?
\par
  \textbf{\mimovlrl{}:} \texttt{<think>...</think>}
To determine the number of cats in the image, we can observe and count:  
- There is a black cat lying on the floor.  
- There is a white and brown cat sitting behind the black cat.  

Thus, the number of cats in the image is 2.  
} \\
\midrule
  \includegraphics[width=7cm]{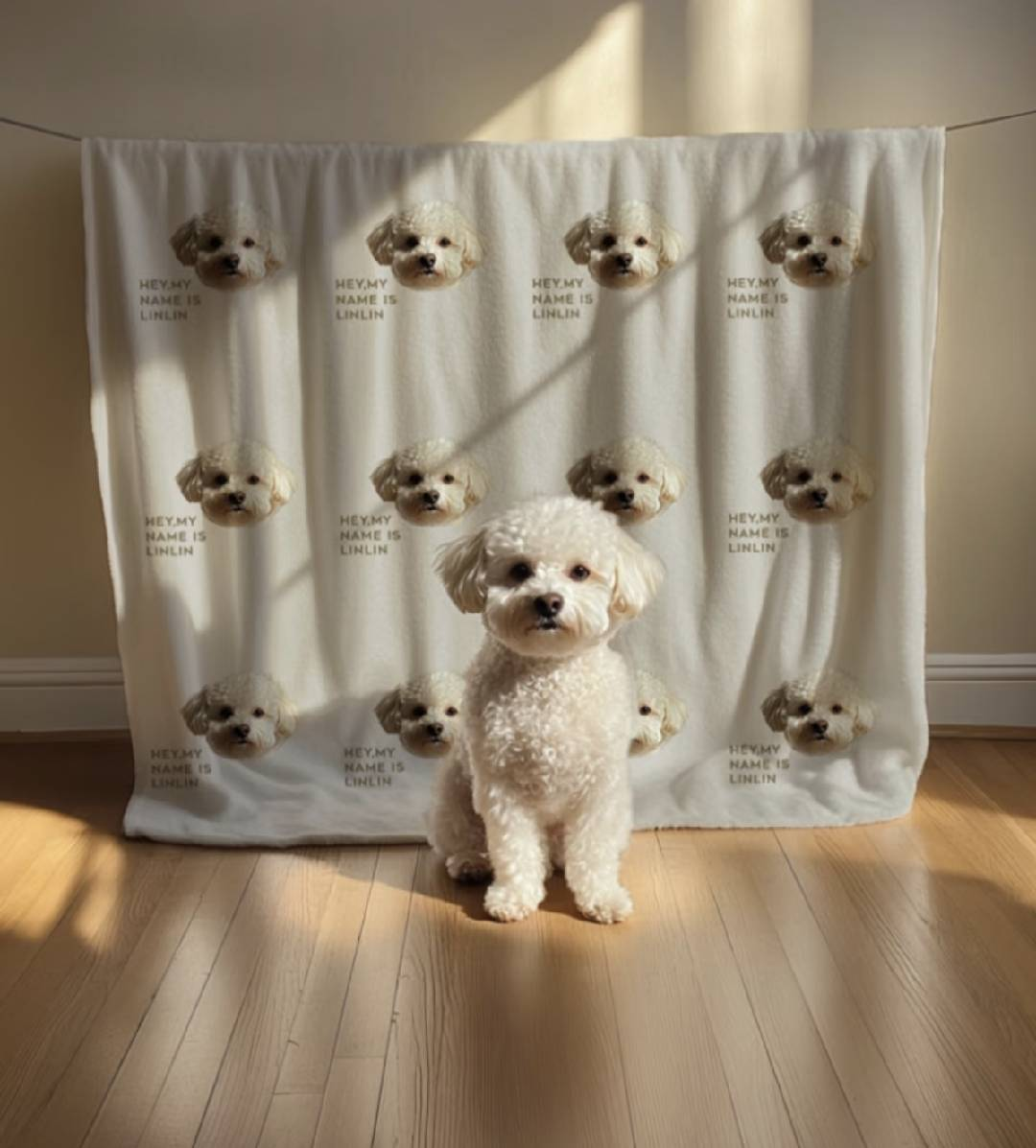} & 
  {\textbf{Instruction:} How many dogs in the image?
\par
  \textbf{\mimovlrl{}:} \texttt{<think>...</think>}
To determine the number of dogs in the image, we analyze both the printed dogs on the blanket and the real dog in the foreground:  
\begin{enumerate}
      \item \textbf{Printed dogs on the blanket}: The blanket features a grid of dog faces. Counting these, we see 3 rows with 4 dogs each, so $ 3 \times 4 = 12 $ printed dogs.  
    \item \textbf{Real dog in the foreground}: There is 1 real dog standing in front of the blanket.    
\end{enumerate}
Adding these together, $ 12 + 1 = 13 $.    

Thus, the total number of dogs in the image is 13.
} \\
  \bottomrule
  \end{tabular}
\caption{Example of \mimovlrl{} solving detailed perception problem.}
\label{fig:case3}
\end{figure}
\renewcommand{\arraystretch}{1}

\clearpage

\renewcommand{\arraystretch}{1.5}
\begin{figure}[h]
  \centering
  \begin{tabular}{m{3cm}m{12cm}}
  \toprule
  \includegraphics[width=3cm]{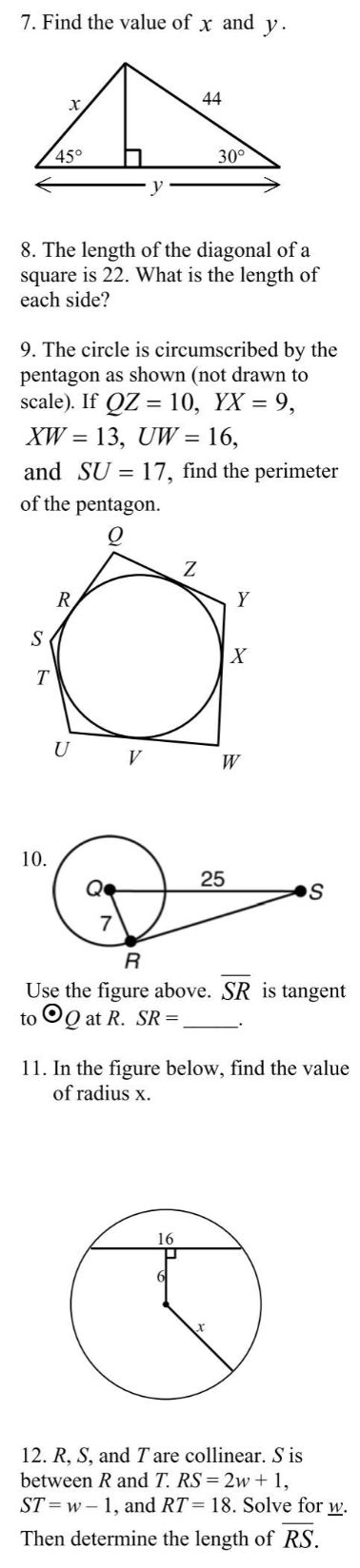} & 
  {\textbf{Instruction:} answer all the questions.\par
  \textbf{\mimovlrl{}:} \texttt{<think>...</think>}
\begin{enumerate}
\item Problem 7  
To find \( x \) and \( y \), we analyze the two right triangles formed by the height:  

- **Right triangle (30° angle)**:  
  Hypotenuse = 44. The height \( h \) (opposite 30°) is \( 44 \cdot \sin(30^\circ) = 44 \cdot 0.5 = 22 \).  
  The base segment \( y \) (adjacent to 30°) is \( 44 \cdot \cos(30^\circ) = 44 \cdot \frac{\sqrt{3}}{2} = 22\sqrt{3} \).  

- **Right triangle (45° angle)**:  
  This is a 45-45-90 triangle, so legs are equal. The height \( h = 22 \), so the other leg (left base segment) is also 22.  
  Hypotenuse \( x \) (45-45-90 ratio: leg \( \cdot \sqrt{2} \)) is \( 22\sqrt{2} \).  

\item Problem 8  
For a square with diagonal \( d = 22 \), the side length \( s \) relates to the diagonal by \( d = s\sqrt{2} \). Solving for \( s \):  
\( s = \frac{d}{\sqrt{2}} = \frac{22}{\sqrt{2}} = 11\sqrt{2} \) (rationalizing the denominator).  

\item Problem 9  
A pentagon circumscribed about a circle (tangential pentagon) has the property that the sum of the lengths of all sides equals \( 2 \times \) the sum of the tangent segments from each vertex. However, since the problem provides the side lengths \( QZ = 10 \), \( YX = 9 \), \( XW = 13 \), \( UW = 16 \), \( SU = 17 \), and the pentagon is tangential, the perimeter is the sum of these sides (as they represent the sides of the pentagon).  

Perimeter \( = 10 + 9 + 13 + 16 + 17 = 65 \).  

\item Problem 10  
\( \overline{SR} \) is tangent to \( \odot Q \) at \( R \), so \( QR \perp SR \) (tangent is perpendicular to radius). Thus, \( \triangle QRS \) is a right triangle with \( QR = 7 \) (radius) and \( QS = 25 \) (hypotenuse). By the Pythagorean theorem:  
\( SR = \sqrt{QS^2 - QR^2} = \sqrt{25^2 - 7^2} = \sqrt{625 - 49} = \sqrt{576} = 24 \)  

\item Problem 11  
A perpendicular from the circle’s center to a chord bisects the chord. The chord length is 16, so half the chord is \( \frac{16}{2} = 8 \). The distance from the center to the chord is 6, forming a right triangle with legs 6 and 8, and hypotenuse \( x \) (radius). By the Pythagorean theorem:  
\( x = \sqrt{6^2 + 8^2} = \sqrt{36 + 64} = \sqrt{100} = 10 \)   

\item Problem 12  
Since \( R, S, T \) are collinear and \( S \) is between \( R \) and \( T \), \( RS + ST = RT \). Substitute \( RS = 2w + 1 \), \( ST = w - 1 \), and \( RT = 18 \):  
\[ (2w + 1) + (w - 1) = 18 \]  
Simplify: \( 3w = 18 \implies w = 6 \).  
Substitute \( w = 6 \) into \( RS \):  
\( RS = 2(6) + 1 = 13 \) 
\end{enumerate}

\textbf{Final Answers}
7. \( x = 22\sqrt{2} \), \( y = 22\sqrt{3} \)  
8. \( 11\sqrt{2} \)  
9. \( 65 \)  
10. \( 24 \)  
11. \( 10 \)  
12. \( w = 6 \), \( \overline{RS} = 13 \)
} \\
  \bottomrule
  \end{tabular}
\caption{Example of \mimovlrl{} solving multiple geometry problems.}
\label{fig:case4}
\end{figure}
\renewcommand{\arraystretch}{1}

\clearpage

\end{document}